\documentclass[paper]{elsarticle}
\usepackage{setspace,listings,color,soul}
\usepackage{lineno,hyperref}
\usepackage{graphicx}
\usepackage{amsmath}
\usepackage{algorithmic}
\usepackage{array}
\usepackage{amssymb}
\usepackage{multirow}
\usepackage{float}
\usepackage{longtable}
\usepackage{cleveref}
\usepackage{graphicx}
\usepackage{caption}
\usepackage{subcaption}

\usepackage{amsthm}
\theoremstyle{plain}

\theoremstyle{plain}
\newtheorem{prop}{Proposition}
\newtheorem{definition}{Definition}
\definecolor{myred}{RGB}{140, 29, 64}
\newcommand\revision[1]{{\color{black}{#1}}}
\newtheorem{remark}{Remark}

\begin{document}

\begin{frontmatter}

\title{Posterior Regularized Bayesian Neural Network Incorporating Soft and Hard Knowledge Constraints}







\address[inst1]{School for Computing and Augmented Intelligence, Arizona State University, Tempe, 85281, AZ, USA}
\address[inst2]{School for Engineering of Matter, Transport and Energy, Arizona State University, Tempe, 85281, AZ, USA}
\author[inst1]{Jiayu Huang}
\author[inst2]{Yutian Pang}
\author[inst2]{Yongming Liu}
\author[inst1]{Hao Yan\corref{mycorrespondingauthor}}
\cortext[mycorrespondingauthor]{Corresponding author.}
\ead{haoyan@asu.edu}

\begin{highlights}
\item We propose a PR-BNN model to incorporate various  knowledge constraints into BNNs.
 
\item Both deterministic and probabilistic views of soft and hard constraints are defined.

\item Efficient algorithms are developed to solve models with soft or hard constraints.

\item Augmented Lagrangian method is applied to solve PR-BNN with multiple hard constraints.

\item In implementations, various knowledge applied in PR-BNN shows advantages over BNNs.
\end{highlights}

\begin{abstract}
Neural Networks (NNs) have been widely {used in supervised learning} due to their ability to model complex nonlinear patterns, often presented in high-dimensional data such as images and text. However, traditional NNs often lack the ability for uncertainty quantification. Bayesian NNs (BNNS) could help measure the uncertainty by considering the distributions of the NN model parameters. 
Besides, domain knowledge is commonly available and could improve the performance of BNNs if it can be appropriately incorporated. In this work, we propose a novel Posterior-Regularized Bayesian Neural Network (PR-BNN) model by incorporating different types of knowledge constraints, such as the soft and hard constraints, as a posterior regularization term. Furthermore, we propose to combine the augmented Lagrangian method and the existing BNN solvers for efficient inference. The experiments in simulation and two case studies about aviation landing prediction and solar energy output prediction have shown the knowledge constraints and the performance improvement of the proposed model over traditional BNNs without the constraints. 
\end{abstract}


\end{frontmatter}


\section{Introduction}
Neural Networks (NNs) have gained great popularity due to their better accuracy in modeling datasets with inherent complex problem structures \cite{dreyfus2005neural}. However, for critical tasks requiring accurate risk quantification in complex engineered systems or applications, where system safety is important, the deterministic prediction model provided by traditional NNs may not be sufficient. 
Recently, combining the power of Bayesian inference and the expressive power of the deep neural networks, Bayesian Neural Networks (BNNs) have been proposed to combine high representation power with the quantifiable uncertainty estimates \cite{specht1990probabilistic}. More specifically, traditional NNs cannot quantify epistemic uncertainty, which refers to the systematic uncertainty due to the uncertainty of the models. BNNs put prior distributions on the Neural Network weights. Furthermore,  the prior distribution of the NN parameters also helps reduce the bias of the model.  Finally, the posterior distribution of the predicted model help quantify the distribution of the output, which could lead to better decision-making in the downstream tasks \citep{krzysztofowicz_bayesian_1999}.  

However, despite the reduction of the bias and good performance in many prediction tasks, purely data-driven BNNs often have the following limitations: 1) Due to the versatility of the Bayesian Neural Network models, the results of the BNN models can still suffer from overfitting due to the complexity of the BNN models, even though the prior regularization is applied. This could be due to the lack of training data or lack of data in a particular testing regime, which may result in poor generalization power.  2) Results from BNN models trained purely based on data may not comply with physical laws or engineering constraints, leading to a large bias or uninterpretable results. 


In engineering systems, complex knowledge constraints may often exist. Common knowledge constraints include the range constraint, smoothness constraint, monotonic relationship between variables, the parametric relationship between variables, etc. Incorporating such general knowledge constraints into the BNNs may be crucial for complex engineering systems to reduce bias and improve interpretability, especially in regions with limited training data. 

In the line of knowledge-constrained learning research, there are two major ways to incorporate constraints into machine learning models, parameter constraints and output constraints. Parameter constraints put the constraints on the model functions to regularize the model training. Parameter constraints have been widely used in linear models, given that these model parameters are often correlated to some physical meanings in engineering systems.  Furthermore, adding such parameter constraints for high-dimensional models is especially important given that it can often avoid overfitting \citep{liao2009learning,feelders2012learning}. These constraints on the model parameters include the smoothness constraints, range constraints \citep{bouzerdoum1993neural}, sparsity constraints \citep{yang2020enhancing}, and low-rank constraints \citep{tai2015convolutional}. Much of the literature focuses on imposing informative prior distribution to empower learning tasks with additional knowledge and constraint in Bayesian inference \citep{dalton2013optimal}. Qualitative knowledge regarding the relationships among variables can be characterized as signs or order constraints, which can be applied via prior construction into Bayesian Network to exhibit more sensible estimates \citep{esfahani_incorporation_2014}. 

However, in complex models such as deep learning models, model parameters often do not have any physical meanings, and specifying a meaningful prior distribution on the model parameters is often impossible. In the literature, another way to integrate knowledge constraints into the machine learning model is the output-based constraint. Instead of adding the constraint on the model parameters, the constraints can be added directly to the output prediction of the machine learning models as the regularization. In literature, Regularized Expected Mean Log-likelihood Prior, Normal Wishart distribution has been proposed as prior to combining some biological knowledge \citep{esfahani_incorporation_2014}. 
However, modifying prior for general knowledge constraints is not straightforward for many different applications due to the black-box models. This is especially true when we have multiple knowledge constraints from different types or data-dependent complex knowledge constraints, and the prior distributions may become harder to define or even not mathematically exist \citep{zhu2014bayesian}. Recently, OC-BNN (Output-Constrained BNN) has been proposed in \cite{yang_incorporating_2021}. However, OC-BNN  work has some limitations on constraint type and prior distribution modification.  
For more details about OC-BNN, please refer to Section \ref{sec:OCBNN}. 

Incorporating domain knowledge via Posterior Regularization (PR) has been proposed into traditional machine learning models \citep{ganchev2010posterior}. PR is defined as a regularized maximum likelihood function, which has the ability to directly constrain the posterior distribution. For more details about the PR framework, please refer to Section \ref{sec:review}.

In this paper, to incorporate more general knowledge constraints in BNNs, we combine the PR framework with the regularized variational inference to create the generic framework named Posterior Regularized BNN (PR-BNN). Our contributions are listed below:
\begin{itemize}
    \item Based on the underlying properties of different knowledge constraints, we define  deterministic and probabilistic scenarios for both soft and hard knowledge constraints.
    \item The proposed PR-BNN applies posterior regularization to incorporate multiple general constraints in the forms of score functions. Both soft and hard constraints can be considered.
    \item The algorithm, with the basis of different Bayesian inference methods, is developed to solve the soft constraint problem under PR-BNN. The efficient algorithm based on the Augmented Lagrangian form is proposed to solve the hard constraint problem.
\end{itemize}

 The rest of this paper is organized as follows. Section 2 is a review of related methods, Section 3 describes the formulation of the Proposed PR-BNN framework, and section 4 and 5 present two simulation experiments and two implementations on real-world data, respectively. Section 6 concludes the paper briefly.

\section{Review of Related Methods}\label{sec:review}
In this section, we would like to briefly introduce some methodologies related to the proposed method. 

\subsection{Review of Bayesian Neural Network (BNNs)}

In this section, the basic BNNs algorithm will be introduced. The notations are listed in Table \ref{BNNs notations}.

\begin{table}[!htbp]
\center
    \caption{Notations in BNNs}
    \label{BNNs notations}
    \begin{tabular}{p{0.15\linewidth}|p{0.6\linewidth}}
        \hline
        \textbf{Notations} & \textbf{Meaning}\\\hline 
        $\mathbf{x}$&  The training inputs\\ \hline
        $\mathbf{y}$ &  The prediction outputs\\  \hline
        $\mathcal{D}$ &  All training variables including input and output variables\\\hline
        $\mathbf{w}$ &  The BNNs parameters such as weights\\ \hline
        $\theta$ &  The parameter vectors for distributions of  $\mathbf{w}$\\ \hline
        $x^*$ &  The new inputs\\ \hline
        $y^*$ &  The predictions corresponding to new input $x^*$\\ \hline
        $q$ & The approximated variational distribution\\
        \hline
    \end{tabular}
\end{table}

Unlike the traditional NNs, model parameters in BNNs $\mathbf{w}$ are distributions, which can be represented as $\mathbf{w}\sim t(\theta)$, where $\theta$ is a deterministic parameter set for the distribution of $\mathbf{w}$. Uncertainty will be invoked by the distribution of $\mathbf{w}$, which is called epistemic uncertainty \cite{hullermeier2021aleatoric} and the over-fitting issue of NNs can be relieved as well. 

Given $n$ input data vectors $\mathbf{x}$ and outputs $\mathbf{y}$, we try to find the posterior distributions of parameters $\mathbf{w}$, that is, $p(\mathbf{w}|\mathbf{x},\mathbf{y})$. Then try to generate the true distribution $p(\mathbf{y}|\mathbf{x})$ via estimated $p(\mathbf{w}|\mathbf{x},\mathbf{y})$.

\begin{equation}
\label{posterior}
\overbrace{p(\mathbf{w}|\mathbf{x},\mathbf{y})}^{\text{Posterior}} = \frac{\overbrace{p(\mathbf{y}|\mathbf{x}, \mathbf{w})}^{\text{Likelihood}}\overbrace{p(\mathbf{w})}^{\text{Prior}}}{\underbrace{p(\mathbf{y}|\mathbf{x})}_{\text{Evidence}}}
\end{equation}

In practice, distribution space of $\mathbf{w}$ is often defined as a prior, denoted as $p(\mathbf{w})$.  $p(\mathbf{y}|\mathbf{x}, \mathbf{w})$ is the likelihood function based on training dataset. The normaliser in Equation ~\ref{posterior}, $p(\mathbf{y}|\mathbf{x})$, is called evidence \citep{gal2016dropout}. Thus, the posterior of $\mathbf{w}$, $p(\mathbf{w}|\mathbf{x},\mathbf{y})$, is evaluated with Equation~\ref{posterior}. 

Unlike the deterministic model, predictive results of BNNs depend on the estimated posterior distribution of $\mathbf{w}$, instead of deterministic $\mathbf{w}$. Predictive distribution will be
\begin{equation}
\label{predictive}
    p(y^*|x^*)=\int p(y^*|x^*,\mathbf{w})p(\mathbf{w}|\mathbf{x},\mathbf{y}) d(\mathbf{w})
\end{equation}

As shown in (\ref{predictive}), the predicted output $y^*$ is the integration over $\mathbf{w}$, which can be also regarded as an expectation formulated as $\mathbb{E}_{p(\mathbf{w}|\mathbf{x},\mathbf{y})}[p(y^*|x^*, \mathbf{w})]$. Unfortunately, this expectation is intractable for any practical case in NNs \citep{blundell2015weight}.

Variational inference \cite{blei2017variational,zhang2018advances} has been widely used to solve the intractability problem in posterior inference of BNNs. Generally, a simpler approximated distribution $q(\mathbf{w}|\theta)$ is used to approximate $p(\mathbf{w}|X,Y)$. Kullback-Leibler (KL) divergence \citep{kullback1951information} is adopted here to measure the distance between distribution $q(\cdot)$ and $p(\cdot)$, which can be denoted as $KL(q(\cdot)||p(\cdot))$. By minimizing  KL divergence over $\theta$, we are able to obtain optimal  variational distribution $q(\cdot)$.

ELBO, a useful lower bound \citep{kingma2013auto} on the negative log-likelihood function,  has been introduced to get the objective function in BNNs as shown in (\ref{hatelbo}) in the training process. 

\begin{equation}
\begin{aligned}
\label{hatelbo}
\widehat{-ELBO}:=\underbrace{KL(q_\theta(\mathbf{w})||p(\mathbf{w}))}_{\text{KL divergence}} \underbrace{-log (p(D|\mathbf{w}))}_{\text{Negative Log-Likelihood (NLL)}}
\end{aligned}
\end{equation}

In the prediction,  Monte Carlo (MC) sampling \cite{hastings1970monte} can be applied to get the mean prediction of output $y$, and  the uncertainty can be estimated as in Equation ~\eqref{mc}. $N$ is the number of MC samplings performed during prediction.

\begin{equation}
\label{mc}
\begin{aligned}
p(y^*|x^*)  &= \mathbb{E}_{p(\mathbf{w}|\mathbf{x}, \mathbf{y})} [p(y^*|x^*, \mathbf{w})]\\
&\approx \frac{1}{N}\sum^N_{i=1}p(y^*|x^*, \hat{\mathbf{w}_i})
\end{aligned}
\end{equation}

Gaining from the NNs structure, BNNs can  provide reliable prediction and uncertainty quantification even for the high-dimensional (HD) data \citep{cabiscol2019understanding}.  Besides, BNNs can quantify complex uncertainty, such as  spatial/temporal variations in HD data, via a rigorous framework \citep{mcdermott_bayesian_2019}.


\subsection{Review of Posterior Regularization (PR)}

In literature, the goal of posterior regularized constraint is to restrict the space of mode posteriors as a way to guide the NN towards the behavior governed by the engineering domain knowledge. Borrowing from the PR framework, we define a general constraint function $f(\mathbf{x, y})$ on output $\mathbf{y}$. It is worth noting that $f(\cdot)$ can be defined on both input $\mathbf{x}$ and output $\mathbf{y}$, therefore, i.e., $f(\mathbf{x},\mathbf{y})$. 

The methods can be represented as follows:
\begin{equation}
\text { PR: } \max _{\Theta} \mathcal{L}(\Theta ; \mathcal{D})+\mathbf{w}(p(\mathbf{y} \mid \mathcal{D}, \Theta)), 
\end{equation}
where $\mathcal{L}(\Theta ; \mathcal{D})$ is the marginal likelihood of $\mathcal{D}$, and $\mathbf{w}(\cdot)$ is a regularization function of the posterior over latent or output variables $\mathbf{Y}$. The posterior here is a posterior for a different learning model instead of the Bayesian posterior. Here, the PR is used to constrain the output of non-Bayesian methods. To generalize PR for the Bayesian method, RegBays \citep{zhu2014bayesian} is proposed by using the variational Bayesian inference form, as shown in Equation \eqref{equ:regbayes}.   

\begin{equation}
\begin{aligned}
\text{ RegBayes: } & \inf_{q(\mathbf{w})}\mathrm{KL}(q(\mathbf{w})\| p(\mathbf{w}|\mathcal{D}))+U(\boldsymbol{\xi})\\
= & \inf_{q(\mathbf{w}),\boldsymbol{\xi}}\mathrm{KL}\left(q(\mathbf{w})\|P(\mathbf{w})\right)-\int_{x}\log P(D|\mathbf{w}))q(\mathbf{w})d\mathbf{w}+U(\boldsymbol{\xi})\\
 & \text{s.t. }q(\mathbf{w})\in\mathcal{P}_{\text{post}}(\boldsymbol{\xi}), 
\end{aligned}
\label{equ:regbayes}
\end{equation}
where $KL(q(\mathbf{w} )\| p(\mathbf{w}\mid\mathcal{D}))$ is the KL-divergence between a desired post-data posterior $q(\mathbf{w} )$ over model weights $\mathbf{w}$ and the true posterior distribution $p(\mathbf{w} \mid \mathcal{D})$. The regularization $\mathbf{U}$ is a function of slack variable $\xi$ (i.e., $\mathbf{U}(\xi) = \|\xi\|_\beta$ or $\mathbf{U}(\xi) = \xi$).  This function provides a flexible way to incorporate additional information, such as domain knowledge. In this work, we will extend RegBayes into a Bayesian deep learning framework to deal with complex data and various types of knowledge constraints and solve the constrained BNNs.





 \subsection{Review of Output-constrained Bayesian Neural Network (OC-BNN)}\label{sec:OCBNN}
Recently, OC-BNN has been proposed to integrate the domain knowledge to the output variable $\mathbf{y}$ of a Bayesian Neural network. The interpretability and generality of the output constraints are not hard to see when we compare them with parameter constraints. More specifically, constraint function $f(\mathbf{y})$ or $f(\mathbf{x},\mathbf{y})$ can be easily defined in the knowledge about the output values or relationship between input and output is given \citep{yang_incorporating_2021} .

 Positive (negative) constraints  $\mathcal{C}^{+(-)}$ are defined as  satisfying that iff $\forall x \in \mathcal{C}_x$, the output $y \in(\notin) {C}_y(x)$. For simplicity, $\mathcal{C}$ represents these two types of constraints in the rest of this section.

 Conditional Output-Constrained Prior $(\boldsymbol{C O C P})$  is proposed by  \cite{yang_incorporating_2021} to incorporate output constraints into prior via Equation \eqref{cocp}:
\begin{equation}
\label{cocp}
p_{\mathcal{C}}(\mathbf{w})=\frac{1}{Z} p(\mathbf{w})\exp(f(\mathbf{y}) )
\end{equation}
 
For multiple constraints, more exponential terms can be multiplied.
The objective function in this method is still variational inference for the BNNs as described in Equation \eqref{hatelbo}. Through modification on prior, the knowledge is incorporated. 

However, the prior distribution constant is often not easy to calculate or even not well defined in many examples. It limits the types of constraints that could be incorporated into the BNNs. For example, the constraint types considered in OC-BNN are mainly about whether the output or the function of output should be in or out of a pre-specified region. Finally, OC-BNN does not have ways to decide the relative weights of different types of constraints, and it doesn't investigate how to guarantee hard knowledge constraints without the use of rejective sampling. 

In our work, we propose a novel knowledge-constrained BNNs framework named Posterior Regularized BNNs (PR-BNNs). We define a more general constraint function format and derive both soft and hard constraints and their corresponding algorithms. PR has been applied to form the objective function to simplify the problem with constraints in the BNN setting. For hard constraints, an Augmented Lagrangian form of the posterior regularized objective function has been derived, and the relative strength of the constraint function can be automatically updated. We discuss the detailed derivation in \Cref{sec: formulation}.

The main differences between OC-BNN work and our work include: First, in OC-BNN work, the constraint type considered is mainly about conditional in or out of region constraints, while in our work, we include more general functions to represent constraints. Also, except for the soft constraints, we define deterministic and probabilistic hard constraints to match  the strictness of knowledge. 
Second, different from OC-BNN, we show that incorporating domain knowledge via posterior regularization, especially in the hard constraint, can lead to a flexible framework that automatically updates the dual variable for each piece of knowledge. This is especially important if we have multiple constraints, given the proposed method can automatically decide the relative importance of each constraint, thereby allowing for robust incorporation. Please see \citep{zhu2014bayesian} paper for more detailed discussions on the advantage of the Posterior Regularization compared to the conditional output regularized prior in the context of traditional machine learning methods.


\section{Formulation\label{sec: formulation}}
\label{section:formulation}
In this section, we will start with the introduction of the knowledge-constraint function in this work, including the definition, types, and examples for different types of knowledge constraints. Then we propose the Posterior-Regularized Bayesian Neural Network (PR-BNN) framework for both soft and hard constraints and efficient algorithms.

\subsection{Definitions of Hard and Soft Knowledge Constraint via PR-BNN}

In this section, we will define two types of constraints that can be used in the PR-BNN, namely, the soft constraint and the hard constraint. 

\begin{definition}
Knowledge Function $f(\mathbf{x},\mathbf{y})$ is a score function of $\mathbf{x} \text{ and } \mathbf{y}$, which is typically defined as the higher, the better (i.e., satisfy the constraint). 
\end{definition}

\begin{remark}
In this paper, we will use the idea of max-margin learning to define the knowledge function $f(\mathbf{x, y})$ by $f(\mathbf{x, y}) = \min (0, s(\mathbf{x, y}) + m)$, which $s(\mathbf{x, y})$ is a score function. By maximizing the knowledge function $f(\mathbf{x, y})$, it will push the knowledge function $f(\mathbf{x, y})$ to $0$ or the score function $s(\mathbf{x, y})$ to be larger than a certain threshold $-m$ for a particular $\mathbf{y}$ given input $\mathbf{x}$ , where m is a very small positive value. 
\end{remark}

Furthermore, we assume that the predicted output $\hat{\mathbf{y}}$ are generated from an NN model $\hat{\mathbf{y}} = g(\mathbf{x},\mathbf{w})$. In BNN, it is commonly assumed that $\mathbf{w}$ is a random variable, which follows a distribution $q(\mathbf{w})$. Therefore, we assume that the randomness of $\mathbf{y}$ only comes from the uncertainty of $\mathbf{w}$. In another word, the set of constraint $f$, $Q=\{E_q [f(\hat{\mathbf{y}})]\geq b\}$ is equivalent on the constraint on $\mathbf{w}$ as $Q_{\mathbf{w}}=\{E_{q(\mathbf{w})} [f(g(\mathbf{x},\mathbf{w}))]\geq b\}$.

 \begin{definition} 
Hard Constraint: \quad  A posterior distribution $q(\mathbf{w})$ is said to satisfy the hard constraint  when $q(\mathbf{w}) \in Q_{\mathbf{w}}=\{E_{q(\mathbf{w})} [f(\mathbf{x, y})]\geq b\}$, where b is a pre-specified constant. Both deterministic and probabilistic hard constraints can be included by the above formula. 
\end{definition}

\begin{prop} Deterministic Hard Constraint: \quad 
If we assume $-c \leq f(\mathbf{x, y})\leq 0$ where $c\geq 0$,  BNN $g(\cdot)$ is a bounded smooth function, and $b = 0$,  $Q_{\mathbf{w}}=\{E_{q(\mathbf{w})} [f(\mathbf{x, y})]\geq 0\}$ since $f(\mathbf{x, y})\leq 0$,  we are able to get $f(g(\mathbf{x},\mathbf{w})) = 0, \quad \forall \mathbf{w}\in q(\mathbf{w})$. 
\end{prop}
This implies that for any posterior output $\mathbf{y}$ generated from the BNN, the constraint should always be satisfied. In this case,  a hard constraint can be defined on the epistemic uncertainty, given all $w$ should make $g(\mathbf{x},\mathbf{w})$ satisfy this constraint.

\begin{definition} 
Probabilistic Hard Constraint ($\epsilon-$satisfied constraint): \quad  We define $f(\mathbf{x, y}) = -I_{(\mathbf{y} \notin C)}$, which $\mathbf{y} \notin C$ implies that $y$ doesn't satisfy the constraint. Therefore, the posterior set with slack variable $\epsilon$ can be interpreted as the $\epsilon-$satisfied constraint as $- P(\mathbf{y}\notin C)+\epsilon= E_q[f(\mathbf{x, y})] +\xi\geq 0$, where $\epsilon\in [0,1]$ is the strictness of the constraint. If $\epsilon>0$, the constraint implies that $ P(\mathbf{y}\notin C)\leq \epsilon.$
\end{definition}

 \begin{definition} 
 Soft Constraint: \quad  A posterior distribution $q(\mathbf{w})$ is said to satisfy the soft constraint defined by the knowledge function $f(\mathbf{x, y})$, when $q(\mathbf{w}) 
 \in Q_{\mathbf{w}}=\{E_{q(\mathbf{w})} [f(\mathbf{x, y})]+ \xi \geq 0\}$, where $\xi$ is a slack variable, instead of a known constant.
 \end{definition} 

If we assume that $\xi$ is known, the method becomes the hard constraint version where $b = -\xi$. However, the soft-constrained version typically assumes that $\xi$ is not known, inspired by the maximum margin learning, we would like to include the slack variable into the loss function.


Examples of $f(\mathbf{x, y})$ formulation for different types of knowledge will be presented in Section \ref{sssec:examples}. Furthermore, we will discuss the algorithm for solving the soft and hard constraints in Section \ref{sec: alg-PRBNN}.

\subsubsection{Examples of Knowledge Constraint Function }\label{sssec:examples}
In this section, we would like to give some examples of the knowledge constraint function on the output variable ($\mathbf{y}$). In order to relax the constraint format, we assume a  score function, which means that if the predicted outputs  accordance  with the knowledge, then the value of $f(\mathbf{x, y})$ would be high, and vice versa.

The types of constraints that can be included in this framework are rather flexible and applicable to different fields. Here, some generic formats of constraints will be listed, and how the design of the constraint functions will be discussed in detail. 

\begin{itemize}
\item \textbf{Conditional value constraint}: Certain transformation functions $r(\cdot)$ of the output values $\mathbf{y}$ should be close to a pre-specified value $C$ as certain independent variables are within an input range $[\mathbf{x}_a,\mathbf{x}_b]$. Then the score function can be defined as $f(\mathbf{x, y})=\min(-|r(\mathbf{x,y})-C|+m,0)$.

\item \textbf{Boundary constraint}: Certain transformation functions $r(\mathbf{x,y})$ should satisfy the bound limit, including the upper and lower bounds (e.g., $[a(\mathbf{x}), b(\mathbf{x})]$). These bounds are not necessarily the same value over a certain range $\mathbf{x}$. Instead, it can be a function of the independent variables. If we have the constraint that $a(\mathbf{x})\leq r(\mathbf{x,y}) \leq b(\mathbf{x})$, the constraint function can be defined as follows: $f(\mathbf{x, y}) = \min(0, -|r(\mathbf{x,y})-\frac{b(\mathbf{x})+a(\mathbf{x})}{2}|)$.  

\item \textbf{Monotonic constraint}: Certain transformation learning $r(\mathbf{x,y})$ on the output $\mathbf{y}$ will result in non-increasing or non-decreasing function. In the time series problem, we can also define some special monotone constraints which allow some small vibration but show some overall trend. For non-decreasing constraint, $f(\mathbf{x, y}) =\min(0,r(\mathbf{x,y})')$.
\item \textbf{Convex and Concave constraint}: Certain transformation learning $r(\mathbf{x,y})$ on the output $\mathbf{y}$ will result in convex and concave functions. In some fields, such as physics, it is common to see some laws including second derivative terms. In that case, we can include this law as a knowledge constraint in BNNs by using the second derivative directly, as follows: $f(\mathbf{x, y})=\min(0,r(\mathbf{x,y})'')$.

\end{itemize}


\begin{remark}
\textbf{Multiple constraints}: 
To incorporate multiple constraints, a linear combination is a simple way to combine several  available and useful constraints with different formats and complexity, as follows:
$$\mathbf{f}(\mathbf{x},\mathbf{y})=\Sigma_i(s_i f_i(\mathbf{x},\mathbf{y})),$$ where $s_i$ is the weight of $i$-th knowledge constraint.
\end{remark}

\subsection{Posterior Regularized Bayesian Neural Networks (PR-BNNs)} \label{sec: alg-PRBNN}

 In this section, we demonstrate core optimization problem solved in proposed PR-BNNs and the corresponding algorithm to solve soft and hard version PR-BNNs, respectively.
 
\subsubsection{Proposed PR-BNNs}
In this work, we focus on the output constraint given any valid input $\mathbf{x}$, whose conditional expectation of the knowledge constraint functions $f(\mathbf{x},\mathbf{y})$, is denoted as $E[f(\mathbf{x},\mathbf{y})\mid \mathcal{D}, \mathbf{w}]$, over the predictive output $\mathbf{y}$ and input set $\mathbf{x}$ using dataset $\mathcal{D}$ and trained weights $\mathbf{w}$. 
For the hard constraint version, the expectation constraint  over the estimated posterior distribution $q(\mathbf{w})$ should be greater than $b$ to satisfy the knowledge constraint:  $E_{q(\mathbf{w})}E[f(\mathbf{x},\mathbf{y})|\mathcal{D},\mathbf{w}] \geq b$. For multiple constraints, we can rewrite constraints into vector form as: $E_{q(\mathbf{w})}E[\mathbf{f}(\mathbf{x},\mathbf{y})|\mathcal{D},\mathbf{w}] \geq \mathbf{b}$.

\paragraph{Hard constrained version}
In other words, by applying this constraint into the variational form of the Bayesian NN, the optimization problem in Equation \eqref{eq:know_con_bnn} is derived.
Bayesian estimation solves
with the constraints applied on the output and inputs of the models as follows:  

\begin{equation}
\begin{aligned}
 & \min_{q(\mathbf{w}),\mathbf{\xi}}KL[q(\mathbf{w})\|P(\mathbf{w})]-E_{q(\mathbf{w})}\log P(\mathbf{y}|\mathcal{D},\mathbf{w})\\
  & s.t. \quad q(\mathbf{w})\in \mathcal{Q},\mathcal{Q}=\{q(\mathbf{w}):E_{q(\mathbf{w})}E[\mathbf{f}(\mathbf{x},\mathbf{y})|\mathcal{D},\mathbf{w}]\geq \mathbf{b}\}, 
\end{aligned}    
\label{eq: hardconst}
\end{equation}
where $\mathcal{Q}$ denotes the regions, where the expected constraint function involving the expected output  is bounded by $b$. 

\paragraph{Soft constrained version}
In the soft-constrained version, it is often useful to allow some violations of the constraints denoted by the slack variable $\xi$. Inspired by the maximum margin learning, we would like to control the amount of violation $\xi$ by a slack penalty $U(\xi)$. 

To generalize the constraints, we would like to introduce the slack variable $\xi$, where the constrained set $\mathcal{Q}_\xi$. To construct the feasible non-empty constraints  $\mathcal{Q}_\xi$ on the posterior distribution estimation $q(\mathbf{w})$, we add a slack penalty $U(\xi)$ to the regularized variational form of the Bayesian form: 
\begin{equation}
\begin{aligned}
 & \min_{q(\mathbf{w}),\mathbf{\xi}}KL[q(\mathbf{w})\|P(\mathbf{w})]-E_{q(\mathbf{w})}\log P(\mathbf{y}|\mathcal{D},\mathbf{w}) + \lambda U(\xi)\\
 & s.t. \quad q(\mathbf{w})\in Q_{\xi},Q_{\xi}=\{q(\mathbf{w}):E_{q(\mathbf{w})}E[\mathbf{f}(\mathbf{x},\mathbf{y})|\mathcal{D},\mathbf{w}]\geq -\xi\}, 
\end{aligned} 
\label{eq: slack}
\end{equation}
where $\mathbf{w}$ is the weights of BNNs; $P(\mathbf{w}|D)$ is the true posterior distribution of $\mathbf{w}$, $q(\mathbf{w})$ is used to estimate the true posterior distribution $P(\cdot)$ due to the intractability of it;

For example, if $U(\xi) = \sum_i \lambda_i \xi_i$,  Equation \eqref{eq: slack} becomes the regularization form
\begin{equation}
  \min_{q(\mathbf{w})}KL[q(\mathbf{w})\|P(\mathbf{w})]-E_{q(\mathbf{w})}\log P(\mathbf{y}|\mathcal{D},\mathbf{w})-\lambda E_{q(\mathbf{w})}E[\mathbf{f}(\mathbf{x},\mathbf{y})|\mathcal{D},\mathbf{w}]  
 \label{equ:new_opt}
\end{equation}
In the following section, we will assume that $U(\xi) = \xi$ in the  algorithm development.

\subsubsection{Optimization Algorithms to Solve PR-BNNs}
In this section, we will derive the optimization algorithms for the PR-BNN. In this paper, we focus on algorithms solving both the hard-constrained version in Equation \eqref{eq: hardconst}  and the soft-constrained version in Equation \eqref{eq: slack}. 

We assume the weights $\mathbf{w}$ in BNNs are distributions with parameter $\theta$, so we need to search for optimal $\theta$ to solve the minimization problem in Equation \eqref{equ:new_opt}.

\paragraph{Soft constrained version}
For the soft constrained version, we will use Equation \eqref{equ:new_opt} as an example, and it is easy to derive that 
\begin{equation}
\begin{aligned} & \arg\min_{\theta}KL[q(\mathbf{w|\theta})\|P(\mathbf{w})]-E_{q(\mathbf{w|\theta})}\log P(\mathbf{y}|\mathcal{D},\mathbf{w})-\lambda E_{q(\mathbf{w|\theta})}E[\mathbf{f}(\mathbf{x},\mathbf{y})|\mathcal{D},\mathbf{w}]\\
= & \arg\min_{\theta}E_{q(\mathbf{w}|\theta)}\mathcal{F}(\mathbf{w},\theta), 
\end{aligned}
\label{eq:know_con_bnn}
\end{equation}
where  $\mathcal{F}(\mathbf{w},\theta) =\log(q(\mathbf{w}|\theta))-\log P(\mathbf{w})-\log P(\mathbf{y}|\mathcal{D},\mathbf{w})-\lambda E[\mathbf{f}(\mathbf{x},\mathbf{y})|\mathcal{D},\mathbf{w}]$. 



\paragraph{Hard constrained version}

For the hard constraint version, we will first introduce the slack variable $z_i$ of knowledge constraint $i$ in order to convert the original inequality constraint into equality constraint. The optimization problem is shown in Equation \eqref{e:hard_obj}
\begin{equation}\label{e:hard_obj}
\begin{aligned}
    &\arg\min_{\theta} \quad KL[q(\mathbf{w}|\theta)\|P(\mathbf{w}|D)],\quad \\ &s.t. \quad E_{q(\mathbf{w}|\theta)}f_{i}(\mathbf{x,y})=z_{i},z_{i}\geq0,i=1,\cdots,I 
\end{aligned}
\end{equation}

The Augmented Lagrangian form can be derived as 
\begin{equation}\label{eq: ALM}
\begin{aligned}
\min_{z, \theta} \quad KL[q(\mathbf{w}|\theta)\|P(\mathbf{w}|D)]-&\sum_{i}s_{i}(E_{q(\mathbf{w}|\theta)}f_{i}(\mathbf{x},\mathbf{y})-z_{i})+\sum_{i}\frac{1}{2}\rho_{i}\|E_{q(\mathbf{w}|\theta)}f_{i}(\mathbf{x},\mathbf{y})-z_{i}\|^{2},\\&s.t. \quad z\geq 0. 
\end{aligned}
\end{equation}

\begin{prop} 
$z_i$ in Equation \eqref{eq: ALM} can be solved as 
\begin{equation}
z_{i}=\max\{E_{q(\mathbf{w}|\theta)}f_{i}(\mathbf{x},\mathbf{y})-\frac{s_{i}}{\rho_{i}},0\}  \label{eq: z}
\end{equation}
\end{prop}
Finally, by plugging in Equation \eqref{eq: z} to Equation \eqref{eq: ALM}, we can derive the following proposition. 
\begin{prop} Solving $\theta$ in Equation \eqref{eq: ALM} is equivalent to solve
\begin{equation}
\min_{\theta} \quad KL[q(\mathbf{w}|\theta)\|P(\mathbf{w}|D)]+\sum_{i}\phi(E_{q(\mathbf{w}|\theta)}f_{i}(\mathbf{w}),s_{i},\rho_{i}),
\label{eq: EQform}
\end{equation}
where

$
\phi(E_{q(\mathbf{w}|\theta)}f_{i}(\mathbf{w}),s_{i},\rho_{i})=\begin{cases}
-s_{i}E_{q(\mathbf{w}|\theta)}f_{i}(\mathbf{w})+\frac{1}{2}\rho_{i}\left(E_{q(\mathbf{w}|\theta)}f_{i}(\mathbf{w})\right)^{2} & E_{q(\mathbf{w}|\theta)}f_{i}(\mathbf{w})\leq s_{i}/\rho\\
-\frac{1}{2}s_{i}^{2}/\rho_{i} & E_{q(\mathbf{w}|\theta)}f_{i}(\mathbf{w})>s_{i}/\rho
\end{cases}.
$
\end{prop}
The proof is similar to  \citep{liu2021novel}. Finally, the dual variable can be updated as  $s_i=\max\{0,s_{i}-\rho_{i}E_{q(\mathbf{w}|\theta)}f_{i}(\mathbf{x},\mathbf{y})\}$ and $\rho_i$ can be updated as $\rho_i = c \rho_i$, where $c$ is typically close to $1$, such as  $c=1.005$. Through the Equation \eqref{eq: EQform}, we are easy to find that the violation of constraint will lead to higher $s_i$, and the violation level also matters.

\subsubsection{BNN Inference}

There are multiple algorithms that can be used for BNN inference efficiently by estimating optimize $\theta$. Overall, these algorithms based on Dropout \citep{gal2016dropout}, Bayes by Back-propagation \citep{blundell2015weight}, Stein Variational Gradient Descent \citep{liu2016stein} and so on. 

\paragraph{Dropout}
By using Dropout layers, nodes will be randomly removed, and the weights of layer $i$, $\mathbf{w}_i$ can be represented as:
$$w_i=M_i*\text{diag}([z_{i,j}]),$$
where $M_i$ is the parameter matrix for layer $i$ and $z_{i,j}\sim \text{Bernoulli}(p_i)$ for each mode $j$. Thus the uncertainty in $w$ comes from the distribution of $z_{i,j}$ for each variational parameter \citep{gal2016dropout}. 

Both Equation \eqref{eq:know_con_bnn}  and  \eqref{eq: EQform}  contains the $E_{q(\mathbf{w}|\theta)}h(\mathbf{w},\theta)$ for some function $h$. It is easy to get sample $w_n$ from Bernoulli distribution $z^n_{i,j}$ and estimate the objective function in Equation \eqref{eq:know_con_bnn} and \eqref{eq: EQform}.

\paragraph{Bayes by Back-propagation}
To optimize $\theta$  in $E_{q(\mathbf{w}|\theta)}h(\mathbf{w},\theta)$ by stochastic gradient descent, $\frac{\partial}{\partial\theta}E_{q(\mathbf{w}|\theta)}h(\mathbf{w},\theta)$ needs to be derived. Here, Bayes by Back-propagation  \citep{blundell2015weight} can be used by utilizing the reparametrization trick. Here, we assume that $\mathbf{w}	\sim N(\mu,\sigma^{2}) \text{, and } \theta=(\mu,\log(\sigma^{2})).$ By the reparameterization trick, $\mathbf{w}$ can be written as $\mathbf{w}	=\mu+\sigma\epsilon,\epsilon\sim N(0,1)$. Then,  $q(\mathbf{w}|\theta)d\mathbf{w}=q(\epsilon)d\epsilon$ can be used to get the gradient more easily. The gradient computation result is  presented  in Equation \eqref{equ:repa}.

\begin{equation}
\frac{\partial}{\partial\theta}E_{q(\mathbf{w}|\theta)}h(\mathbf{w},\theta)=E_{q(\epsilon)}[\frac{\partial h(\mathbf{w},\theta)}{\partial\mathbf{w}}\frac{\partial\mathbf{w}}{\partial\theta}+\frac{\partial h(\mathbf{w},\theta)}{\partial\theta}]
\label{equ:repa}
\end{equation}

\paragraph{Stein Variational Gradient Descent}

By defining a Stein operator and stein discrepancy, we can avoid calculating the normalization constant in Bayesian rules and get the derivative of the objective function more easily \citep{liu2016stein}.
The gradient of the objective function will only depend on 

\begin{equation}
\begin{aligned}
\nabla (\log (P(\mathbf{w}|\theta))-\lambda \mathbf{f}(\mathbf{x},\mathbf{y}))
  &=\nabla (\log (P(D|\mathbf{w})P(\mathbf{w}))-\lambda \mathbf{f}(\mathbf{x},\mathbf{y}))\\
  &=\nabla \log P(D|\mathbf{w})+ \nabla \log P(\mathbf{w})-\lambda \nabla \mathbf{f}(\mathbf{x},\mathbf{y}). 
\end{aligned}
\label{equ:svgd}
\end{equation}

The initial set of particles $w^0$ are sampled from the initial estimated posterior distribution $q^0$. Through updating particles iteratively, we can finally obtain a set of particles $w$ to match target distribution $p(\mathbf{w})$.


Our proposed PR-BNNs can be solved via different bayesian inference method as shown above. The complexity of the inference and computation depends on the Bayesian inference method selected.

\section{Implementation}
In this section, we will evaluate the methods using some simulation examples. The behavior of the algorithms in these simple examples would be helpful in understanding the behavior of the proposed algorithm. More specifically, we will evaluate the case of soft constraint and hard constraint in Section \ref{subsec: soft} and Section \ref{subsec: hard}.

\subsection{Simulation 1: Soft constraint} \label{subsec: soft}
 
\begin{figure}
 \centering
  \centering
 \includegraphics[scale=0.18]{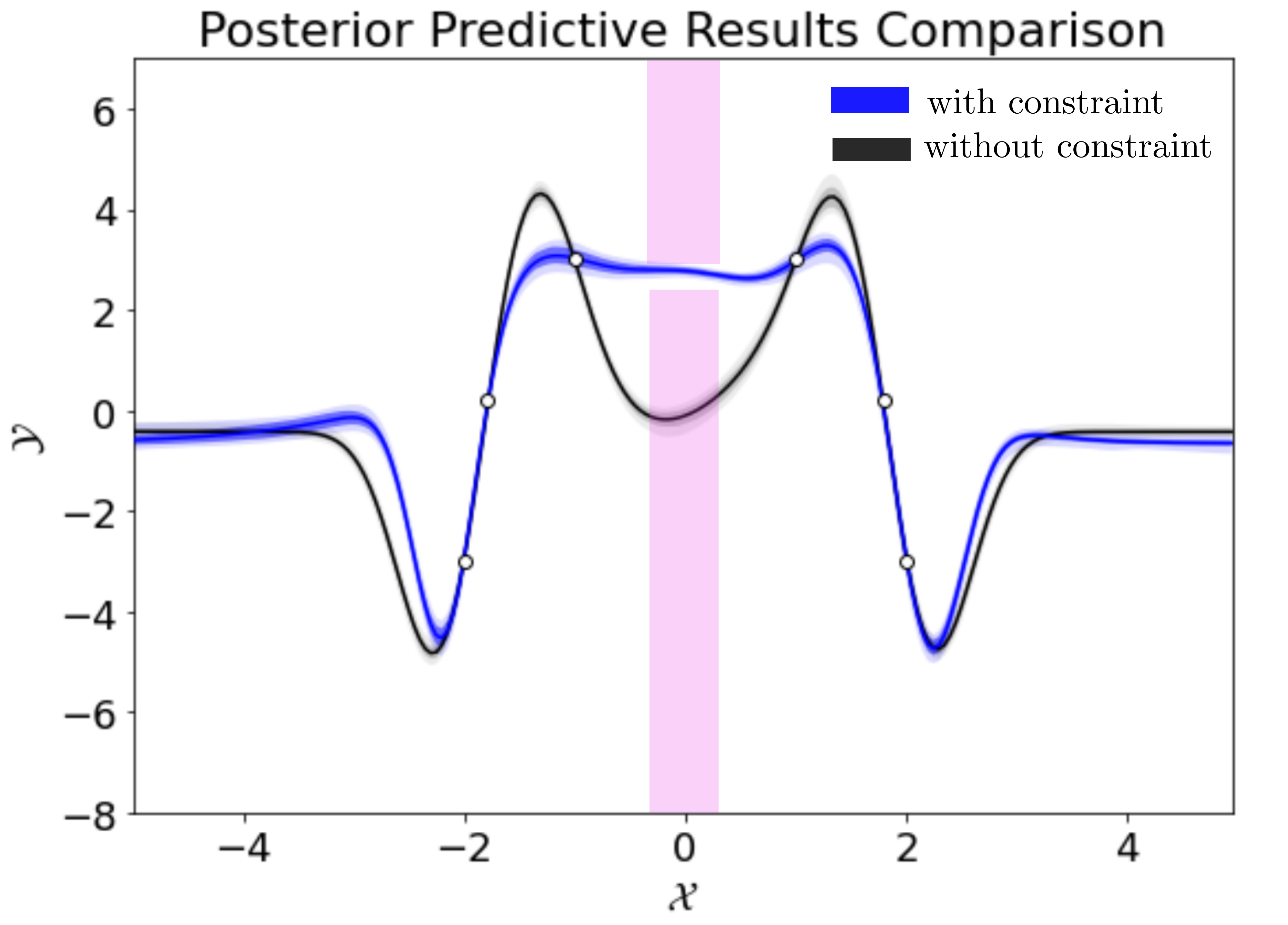}
 \caption{Results generated from different Bayesian inference methods}
 \label{fig:ocbnn}
 \end{figure}

In this experiment, the same BNNs architecture and example in  \citep{yang_incorporating_2021} is used. The BNNs consist of a single 10-node RBF hidden layer and a Radial basis function (RBF) as an activation function. For RBF activation function, given an input $x$, the output after activation could be $\Phi (x) =  e^{- \frac{(x-\mu)^2}{\sigma^2}}$. For the prior, the paper suggests  an isotropic Gaussian $\mathbf{
w} \sim \mathcal{N}\left(\mathbf{0}, \sigma_{\mathbf{w}}^{2} \mathbf{I}\right)$ that is first proposed by MacKay \citep{mackay1995probable} due to the tractability. $\sigma_{\mathbf{w}}^{2}=1$ is the shared variance for weights $\mathbf{w}$. We have six data points in the training set. In this experiment, it is hypothesized that the Knowledge-constrained model can learn those given side information or knowledge. Bayes by Back-propagation (BBB) \citep{blundell2015weight} is used in this study.
The constraint is in the same format as in  \citep{yang_incorporating_2021}: $y\in [2.5,3]$ when $\text{for }x\in [-0.3,0.3]$. After adding this conditional range constraint, the posterior predictive results $y_s$ are within $[2.5,3]$ as the blue curve in Figure \ref{fig:ocbnn}. Without this constraint, the predictive results represented in black in Figure \ref{fig:ocbnn} will not follow the conditional constraint. 

\subsection{Simulation 2: Hard constraint} \label{subsec: hard}

\begin{figure}[h]
\centering
\includegraphics[width=0.99\columnwidth]{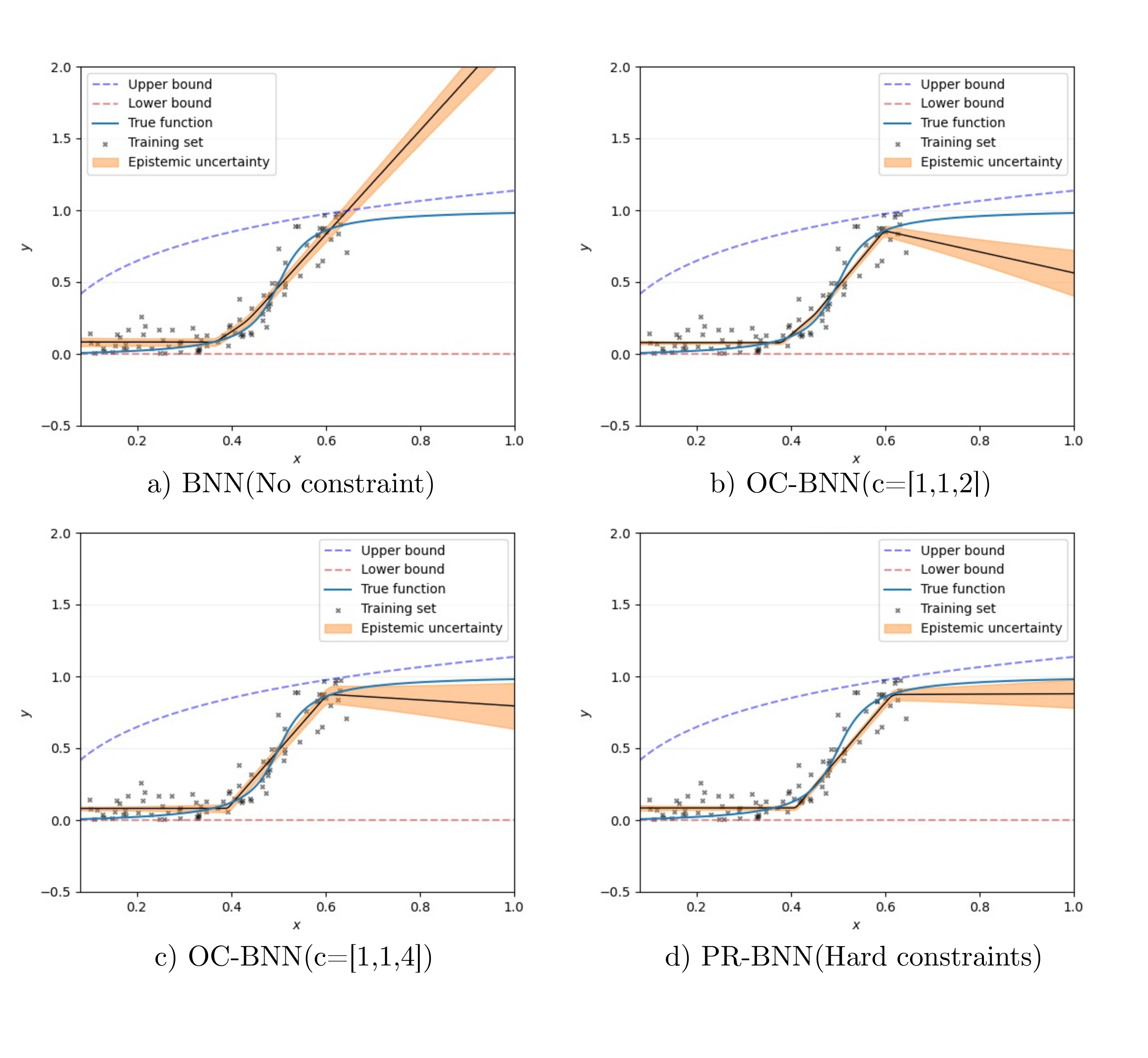}
\caption{Experiment details and results comparison}
\label{fig:simple_sim}
\end{figure}

\paragraph{Simulation Setup} As mentioned, incorporating knowledge constraints may help improve the model performance  when we have noisy or limited training data.  In a simple regression simulation, the training data is noisy and insufficient as shown in Figure \ref{fig:simple_sim}(a).  In testing, we have data points of $x$ values between $0.08$ and $1$, while data points available for training are of $x$ values only between $0.1$ and $0.65$. The true function is $y=(\arctan(20x - 10) - \arctan(-10))/3, $ for $x\in [0.08,1]$ as the blue curve shown in Figure \ref{fig:simple_sim}(a)  \citep{agrell_gaussian_2019} . 

For the constraints, we use: (i) boundary constraint: $a(x) \leq r(x) \leq b(x)$.  (ii) non-decreasing constraint.
\begin{itemize}
    \item \textbf{Constraint 1: lower bound constraint}: $y \geq a(x)=0$ is the lower bound, as shown in red line of dashes shown in Figure \ref{fig:simple_sim}. \revision{The constraint function can be presented as $f_1(y)=\min\{0,y\}$;}
    \item \textbf{Constraint 2: upper bound constraint}: $y \leq b(x) = (\log(25x + 1))/3+0.05$ is the upper bound, as shown in the purple line of dashes. \revision{The constraint function is $f_2(y)=\min\{0,b(x)-y\}$;}
    \item \textbf{Constraint 3: monotone constraint}: $\partial f / \partial x_{i} \geq 0$. \revision{The constraint function is $f_3(y)=\min\{0,y^\prime\}$;}
\end{itemize}

\revision{\paragraph{Methods for Evaluation} 

Here, we will compare the performance of proposed PR-BNN and two benchmark models: OC-BNN and Baseline-BNN. More details of each method are presented here:
\begin{itemize}
    \item \textbf{Proposed PR-BNN}: We apply a BNN with two fully-connected Bayesian linear layers to fit the 1D function using data points in Figure \ref{fig:simple_sim}(a). The first layer has 100 nodes followed by a ReLU activation function and the second layer is output layer with only 1 node. Besides, we add and update aleatoric uncertainty to realize automatic parameter tuning for traditional error and complexity loss.
    \item \textbf{Baseline-BNN}: We use the exact Bayesian Neural Network architecture in the PR-BNN but without adding the constraints. 
    \item \textbf{OC-BNN}: OC-BNN only supports limited types of constraints, such as positive and negative constraints, where monotone constraints are not directly supported. Furthermore, there is no guidance on how to set appropriate weights for multiple constraint types. Therefore, we have made a modification to convert the constraint on $y$ to the positive constraint on the first-order gradient of $y$, denoted as $y' \in [0,\infty)$. In this case, OC-BNN can be extended and solve both boundary and monotone constraints. Finally, we follow the definition of the paper to provide a simple way to incorporate multiple constraints in the modified prior, including constraints as shown in Equation \eqref{cocp}.

For the case with multiple constraints, we can use rewrite Equation \eqref{cocp} as 
$$
p_{\mathcal{C}}(\mathbf{w})=\frac{1}{Z} p(\mathbf{w})\exp(\Sigma_i( c_i*f_i(\mathbf{y})) )=\frac{1}{Z} p(\mathbf{w})\prod \exp(c_i*f_i(\mathbf{y})),
$$
where $c_i$ is a hyperparameter for output constraint $f_i(\mathbf{y})$.

Since they haven't proposed an efficient algorithm for hard constraints, we treat the aforementioned three constraints as soft ones by introducing the corresponding weight $c_i$. Parameter tuning is challenging, given that we have many constraints with a lot of combinations of their weights. Let the vector $c=[c_1, c_2, c_3]$ demotes weights for constraint 1, 2 and 3 respectively. To implement the OC-BNN work into this simulation case, we use multiple parameter combinations of $c$ and record results. 
\end{itemize}

\paragraph{Results} 
To evaluate the performance, the following evaluation metrics are used 1) \textit{Mean Squared Error (MSE)}: MSE is used to evaluate the accuracy of mean prediction.  2) \textit{Epistemic Uncertainty Standard deviation (STD)}: Epistemic uncertainty is computed by the standard deviation of the Monte Carlo sampling of all BNN methods averaged over the input domain.  3) \textit{Violation value for constraint $i$ as $v_i$}: $v_i$ is computed by the expected value of the constraint function as $v_i=-E_w(f_i(y))$, where $f_i$ is the constrained function. 4) \textit{Number of Violations for constraint $i$ as $n_i$}: $n_i$ shows the number of constraint $i$ violation predictions among 200 testing samples. The results of these four groups of metrics of PR-BNN, OC-BNN, and BNN are shown in Table \ref{t:s2ocbnn_violation}. 

\revision{For PR-BNN with hard constraints, all three constraints have been satisfied, and the detailed results for each constraint are shown in the last row of Table \ref{t:s2ocbnn_violation}. The dual variables $s_i$ can be considered as the weights of corresponding constraints, which are automatically decided by the augmented Lagrangian algorithm. The initial setting is $s=[1,1,1]$, and after training, final vector $s$ is  $[4.36,14.41,6.41]$. Their rounded numbers are shown in the last row of Table \ref{t:s2ocbnn_violation}. We can observe that $s_2=14.41$ is far greater than the other two dual variable values, indicating that the upper bound constraint violation is more severe in the training stages and large $s_2$ accelerate the process of learning this constraint.}

Comparing PR-BNN and BNN, MSE is much smaller for the proposed PR-BNN method, given that constraints provide useful information, especially for the range without any data points. Without constraints, a severe violation of the upper bound is shown in Figure \ref{fig:simple_sim}(a), while most predictions and mean predictions satisfy the bound constraint after constraints injection as shown in \ref{fig:simple_sim}(d).  The MSE loss is reduced from 0.2320 in the BNN model to 0.0052 in the PR-BNN model after adding constraints. The epistemic uncertainty is reduced from 0.059 in the unconstrained model to 0.040 in the hard constrained model, benefiting from the knowledge, and we can easy to observe this uncertainty reduction through Figure \ref{fig:simple_sim}(a) and (d). 

For OC-BNN, we would like to evaluate different sets of weight coefficients to combine the constraints. From Table \ref{t:s2ocbnn_violation}, we can conclude that the upper bound (Constraint 2) and Monotone constraint (Constraint 3) are competing during training. Due to the space limit, we only include the visualization for hyperparameter settings $c=[1,1,2] \text{ and } c=[1,1,4]$ in Figure \ref{fig:simple_sim}(b) and Figure \ref{fig:simple_sim}(c) respectively  but constraints violation in Table \ref{t:s2ocbnn_violation}  shows details for more experiments, where $v_i$ represents the violation value for constraint $i$. From the visualization samples in Figure \ref{fig:simple_sim}(b), we can observe that even though we add double weights on the monotone constraint (Constraint 3), the upper bound constraint (Constraint 2) still pushes the predicted curve too low to violate the monotone constraint. Thus, when we have multiple constraints, it is hard to satisfy all of them via the OC-BNN without time-consuming parameter tuning, and hard to find a balanced setting. As shown in Figure \ref{fig:simple_sim}(c), the prediction becomes better when we add more weights on Constraint 3, but a very small violation value is still shown in the corresponding row of Table \ref{t:s2ocbnn_violation}. Besides, under the $[1, 1, 8]$ parameter setting, the upper bound will be violated greatly since the weight of the monotone constraint is too high. From the comparison in Table \ref{t:s2ocbnn_violation} we can see the proposed PR-BNN can satisfy all the hard constraints without manually parameter tuning.


\begin{table}[h]
\centering
\caption{\revision{Performance comparison on testing set} }
\revision{\begin{tabular}{|c|c|c|c|c|c|c|c|c|c|}
\hline
Framework                                                                                                                & c/s           & $v_1$ & $v_3$    &$v_3$       & $n_1$ & $n_2$ & $n_3$  & MSE     & STD   \\ \hline
BNN                                                                                                        & -           & 0  & 0.22 & 0       & 0  & 77 & 0  & 2.3e-1 & 0.059 \\ \hline
\multirow{6}{*}{\begin{tabular}[c]{@{}c@{}}OCBNN \\ with \\ different  \\ parameter \\vector \\ (c) \end{tabular}} & {[}1,1,1{]} & 0  & 0    & 2.0e-3 & 0  & 0  & 84 & 3.6e-2 & 0.047 \\ \cline{2-10} 
                                                                                                           & {[}1,1,2{]} & 0  & 0    & 1.4e-3 & 0  & 0  & 86 & 3.2e-2 & 0.047 \\ \cline{2-10} 
                                                                                                           & {[}1,2,1{]} & 0  & 0    & 2.7e-3 & 0  & 0  & 84 & 5.9e-2 & 0.046 \\ \cline{2-10} 
                                                                                                           & {[}2,1,1{]} & 0  & 0    & 1.7e-3 & 0  & 0  & 84 & 3.3e-2 & 0.052 \\ \cline{2-10} 
                                                                                                           & {[}1,1,4{]} & 0  & 0    & 3.9e-4 & 0  & 0  & 83 & 8.3e-3 & 0.060 \\ \cline{2-10} 
                                                                                                           & {[}1,1,8{]} & 0  & 0.11 & 0       & 0  & 67 & 0  & 8.6e-2 & 0.079 \\ \hline
PR--BNN (s)                                                                                                  & {[}4,14,6{]}           & 0  & 0    & 0       & 0  & 0  & 0  & 5.2e-3 & 0.040 \\ \hline
\end{tabular}
}
\label{t:s2ocbnn_violation}
\end{table}

}

\section{Case studies}

In this section, we will evaluate the proposed methods in two case studies. First, we will investigate the effect of the knowledge constraints added in the Aviation Landing Status Prediction in Section \ref{subsec: aviation}. Second, we will evaluate how such constraints can be added to solar power prediction using the knowledge from the inverter machine in Section \ref{subsec: solar}. 

\subsection{Aviation Landing Trajectory Prediction \label{subsec: aviation}}
In air traffic management and safety control, aircraft landing is one of the riskiest phases of flight with multiple possible adversities, such as misalignment, hard landing, runway overrun, etc \citep{wang2019aircraft}. Therefore, in this work, we focus on predicting the aircraft landing tracks with real-world aviation radar recordings. We obtain the flight track data from the Sherlock Data Warehouse \citep{arneson2019sherlock}. Sherlock is a platform to collect, archive, process, and query aviation datasets such that to support ATM research \citep{pang2022bayesian}. It contains aviation flight data, meteorological data, and airline operations data from various sources (e.g., FAA and NOAA). The flight tracks are the time-series aircraft coordinates landed at runway 26R at KATL (Hartsfield - Jackson Atlanta International Airport) from January 1st, 2019 to January 31st, 2019, for the case studies. We perform data preprocessing to filter out the last several timestamps since the tracks from Sherlock are the recordings for the entire flight from take-off to landing. The preprocessing of the landing tracks can be summarized as follows: a) perform altitude filtering to determine the touchdown point (landing point on the runway) from the time series; b) conduct runway checks to make sure the given track landed at the correct runway; c) do completion checks to make sure the processed flight tracks are valid and only including a fix time horizon prior to touchdown. Upon finishing data filtering and preprocessing, the dataset used for model training, validation, and testing consists of 7,542 landing trajectories with a fixed horizon of 50 timestamps before the touchdown. 

\begin{figure}[!t]
    \centering
    \includegraphics[scale=0.9]{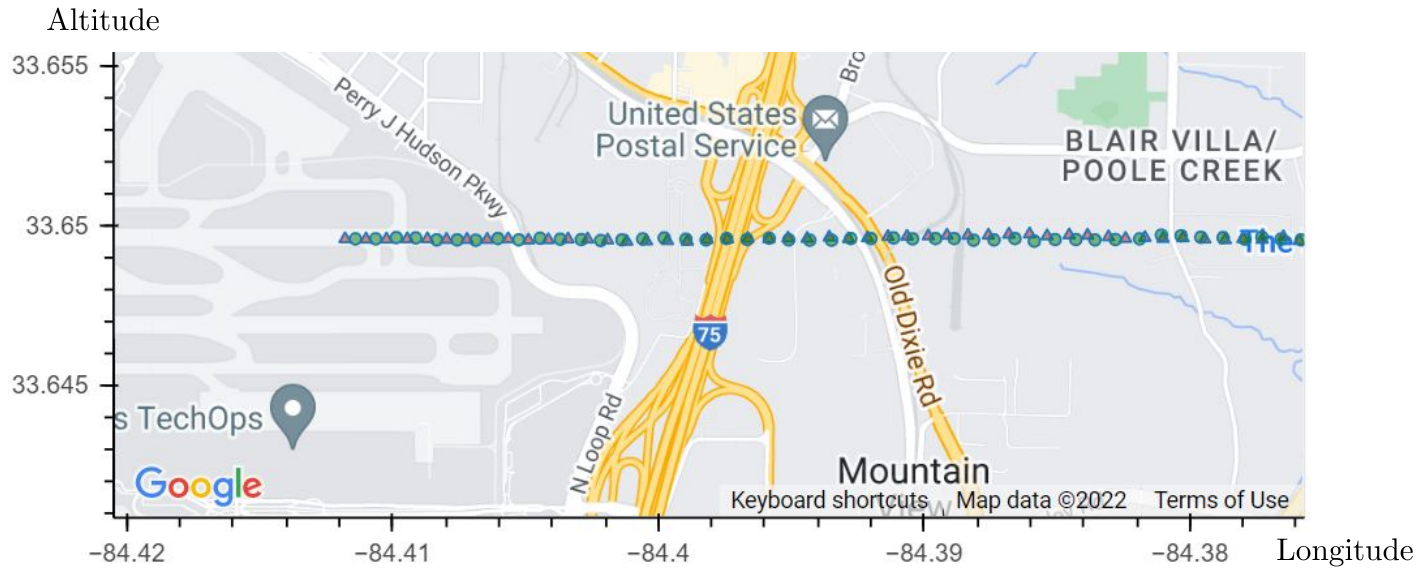}
    \caption{Landing Trajectory Examples}
    \label{fig:traj}
\end{figure}

Here, 5 different variables are collected to picture the flight status in the final 50 seconds before touchdown. Altitude, latitude, and longitude can describe the location of the airplane and we have 2-D landing trajectory sequence examples including longitude and latitude for two flights shown in Figure \ref{fig:traj}; airspeed and course can characterize the speed of the airplane relative to the air mass \citep{gracey1980measurement} and heading direction, respectively. Data for all variables is normalized using Min-Max scalar. Without loss of generality, we use  $x$ to represent any variable in this dataset and the normalization can be shown in Equation \eqref{e:scaler}.

\begin{equation}
\label{e:scaler}
x_{min\_max}=\frac{x-x_{min}}{x_{max}-x_{min}}
\end{equation}

However, the dataset generated by multiple sensors from airports and aircraft is rather noisy, which troubles accurate prediction during the landing phase. 
As a safety-critical problem, landing speed forecasting and uncertainty quantification are of great importance for landing risk mitigation. Integrating common knowledge in BNNs in the model of the landing phase helps reduce prediction uncertainty and improve safety. Furthermore, accurate prediction of the normal landing trajectory with the knowledge-constrained BNN prediction can also help to detect anomalies in the landing phase that seriously violate such knowledge constraints and send early warnings to control tower.



 \begin{figure}[!t]
    \centering
    \includegraphics[scale=0.2]{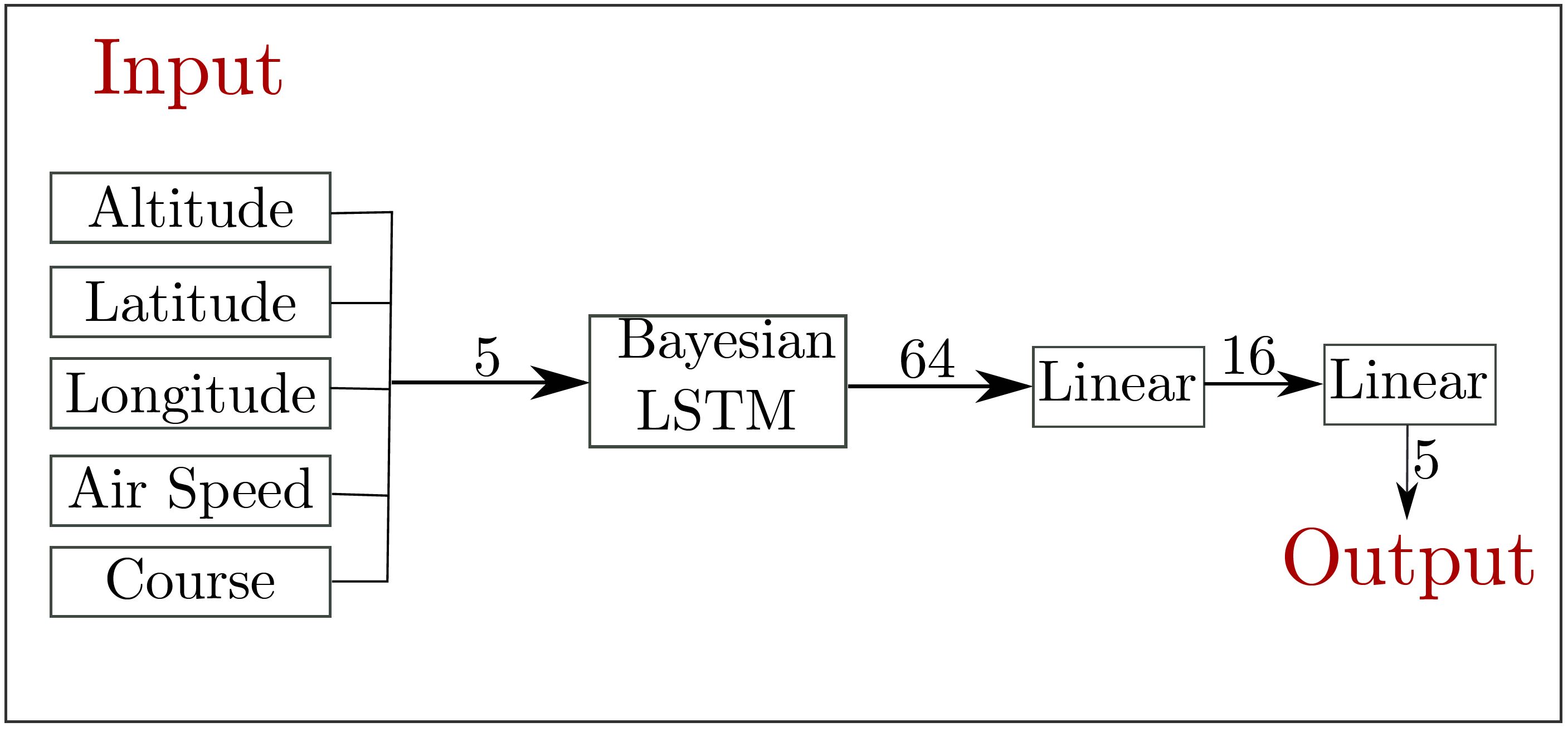}
    \caption{Bayesian LSTM Structure}
    \label{fig:lstm_structure}
\end{figure}

 

Based on this dataset, we would use the first 35 seconds of data to predict the flight status of the last 15 seconds before landing touchdown. A long short-term memory model (LSTM) is used to model this time series data and via Bayesian LSTM as described in  \citep{esposito2020blitzbdl}, with modifications to incorporate constraints. 

The Bayesian LSTM model used in this case study consists of one Bayesian LSTM layer and simply two fully connected layers.

We have 4 different constraints in this case:
\begin{itemize}
    \item Entire Trajectory: Altitude decreasing constrain. In practice, we allow altitude fluctuation. The implementation of this constraint is described as: In the last 15 seconds, the altitude at $t_i$ should greater than the altitude at $t_{i+3}$.
    \item Entire Trajectory: Range constraints on all five variables; For speed and altitude, they should be non-negative. For course, longitude and latitude, range constraints can be set based on the location and direction of the runway.
    \item Landing Point Constraint: Altitude range constraint in landing; Altitude range will be very narrow in touchdown moment due to the narrow change in altitudes of the runway. 
    \item Landing Point Constraint: Speed range constraint in landing; The touchdown airspeed should be within a reasonable and safe range.
\end{itemize}

Here, we have added the constraints as the soft constraints, given that there can actually be very  rare violations of these constraints during some uncontrollable rare cases. In this case, $\xi$ in Equation \eqref{eq: slack} is a slack variable, instead of a known constant.

To show the performance of the PR-BNN model, we compare the 15-second sequence prediction and last point prediction before and after adding constraints. The Mean Squared Error (MSE), the uncertainty which is represented by Standard Deviation (STD) and The continuous ranked probability score (CRPS) a   performance measurement for probabilistic forecasts is shown in Table \ref{t:mse}, Table \ref{t:std}, and Table \ref{t:crps}, respectively.

\begin{table}[!h]
\centering
\caption{Mean squared error results comparison}
\begin{tabular}{|c|c|c|c|c|}
\hline
Model      & \multicolumn{2}{c|}{MSE without   Constraints}               & \multicolumn{2}{c|}{MSE with Constraints}                    \\ \hline
Variables  & \multicolumn{1}{c|}{sequence}           & last point         & \multicolumn{1}{c|}{sequence}           & last point         \\ \hline
latitude   & \multicolumn{1}{c|}{\textbf{1.806E-04}} & \textbf{1.500E-04} & \multicolumn{1}{c|}{1.813E-04}          & 1.514E-04          \\ \hline
longitude & \multicolumn{1}{c|}{9.160E-05}          & 1.838E-04          & \multicolumn{1}{c|}{\textbf{7.050E-05}} & \textbf{9.710E-05} \\ \hline
altitude   & \multicolumn{1}{c|}{3.079E-04}          & 5.186E-04          & \multicolumn{1}{c|}{\textbf{2.450E-04}} & \textbf{2.655E-04} \\ \hline
speed      & \multicolumn{1}{c|}{1.773E-04}          & 1.870E-04          & \multicolumn{1}{c|}{\textbf{1.709E-04}} & \textbf{1.813E-04} \\ \hline
course     & \multicolumn{1}{c|}{\textbf{7.563E-04}} & \textbf{5.462E-04} & \multicolumn{1}{c|}{7.591E-04}          & 5.477E-04          \\ \hline
overall    & \multicolumn{1}{c|}{3.027E-04}          & 3.171E-04          & \multicolumn{1}{c|}{\textbf{2.853E-04}} & \textbf{2.486E-04} \\ \hline
\end{tabular}
\label{t:mse}
\end{table}

\begin{table}[!h]
\centering
\caption{Standard deviation results comparison}
\begin{tabular}{|c|c|c|c|c|}
\hline
Model      & \multicolumn{2}{c|}{STD without   Constraints} & \multicolumn{2}{c|}{STD with Constraints}                    \\ \hline
Variables  & \multicolumn{1}{c|}{sequence}    & last point  & \multicolumn{1}{c|}{sequence}           & last point         \\ \hline
latitude   & \multicolumn{1}{c|}{2.907E-03}   & 2.760E-03   & \multicolumn{1}{c|}{\textbf{2.411E-03}} & \textbf{2.379E-03} \\ \hline
longitude & \multicolumn{1}{c|}{8.241E-03}   & 1.066E-02   & \multicolumn{1}{c|}{\textbf{5.913E-03}} & \textbf{4.854E-03} \\ \hline
altitude   & \multicolumn{1}{c|}{8.149E-03}   & 1.178E-02   & \multicolumn{1}{c|}{\textbf{5.182E-03}} & \textbf{4.209E-03} \\ \hline
speed      & \multicolumn{1}{c|}{3.983E-03}   & 4.235E-03   & \multicolumn{1}{c|}{\textbf{3.255E-03}} & \textbf{3.150E-03} \\ \hline
course     & \multicolumn{1}{c|}{1.676E-03}   & 1.554E-03   & \multicolumn{1}{c|}{\textbf{1.283E-03}} & \textbf{1.297E-03} \\ \hline
overall    & \multicolumn{1}{c|}{4.991E-03}   & 6.198E-03   & \multicolumn{1}{c|}{\textbf{3.609E-03}} & \textbf{3.178E-03} \\ \hline
\end{tabular}
\label{t:std}
\end{table}

\begin{table}[!h]
\centering
\caption{CRPS results comparison}
\begin{tabular}{|c|cc|cc|}
\hline
Model     & \multicolumn{2}{c|}{CRPS without   Constraints}      & \multicolumn{2}{c|}{CRPS with Constraints}                    \\ \hline
Variables & \multicolumn{1}{c|}{sequence}  & last point         & \multicolumn{1}{c|}{sequence}           & last point         \\ \hline
latitude  & \multicolumn{1}{c|}{1.091E-02} & 8.628E-03          & \multicolumn{1}{c|}{\textbf{9.305E-03}} & \textbf{8.451E-03} \\ \hline
longitude & \multicolumn{1}{c|}{1.848E-02} & 1.641E-02          & \multicolumn{1}{c|}{\textbf{5.203E-03}} & \textbf{5.990E-03} \\ \hline
altitude  & \multicolumn{1}{c|}{1.841E-02} & 1.442E-02          & \multicolumn{1}{c|}{\textbf{1.008E-02}} & \textbf{1.064E-02} \\ \hline
speed     & \multicolumn{1}{c|}{2.180E-02} & 2.130E-02          & \multicolumn{1}{c|}{\textbf{8.682E-03}} & \textbf{8.263E-03} \\ \hline
course    & \multicolumn{1}{c|}{2.043E-02} & \textbf{1.670E-02} & \multicolumn{1}{c|}{\textbf{2.035E-02}} & 1.686E-02          \\ \hline
overall   & \multicolumn{1}{c|}{1.801E-02} & 1.549E-02          & \multicolumn{1}{c|}{\textbf{1.072E-02}} & \textbf{1.004E-02} \\ \hline
\end{tabular}
\label{t:crps}
\end{table}

\begin{figure}[!t]
\centering
\includegraphics[width=.85\textwidth]{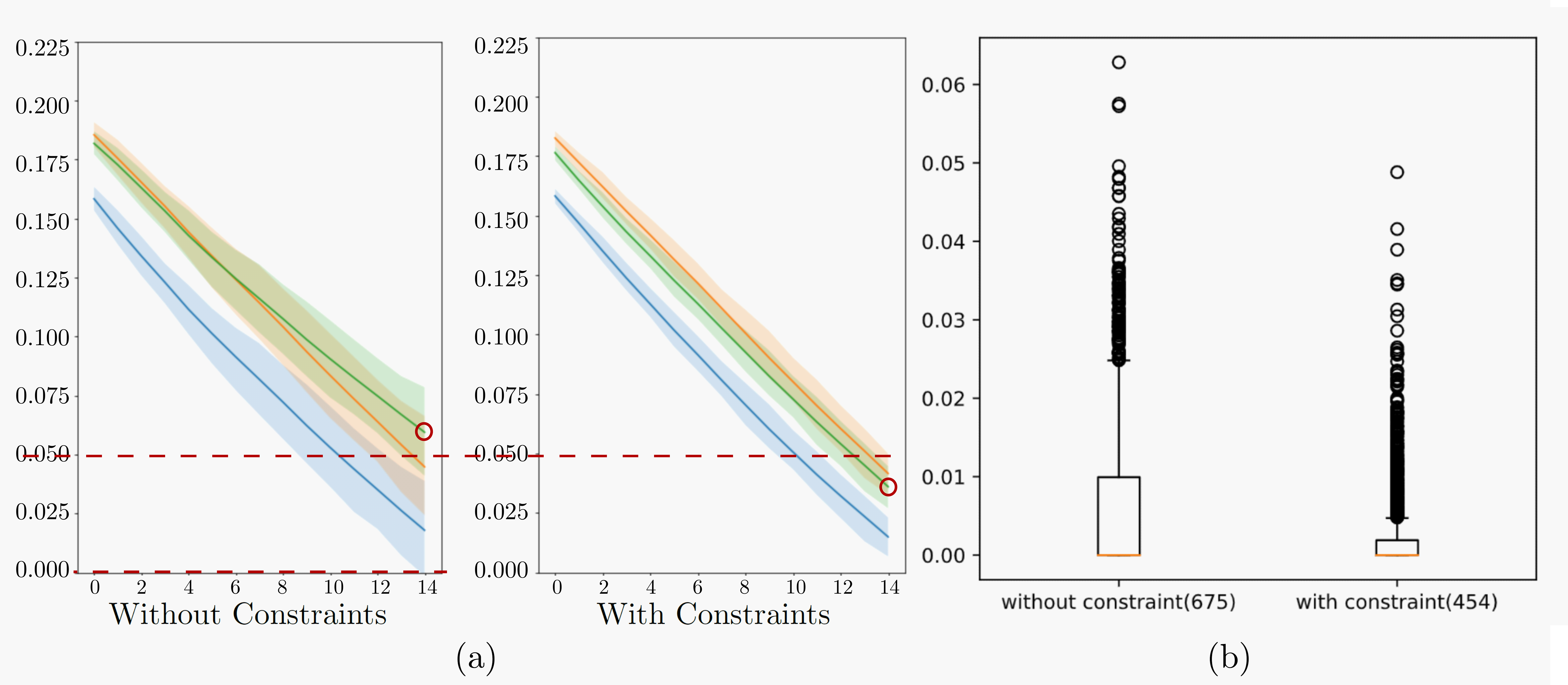}
\caption{(a): Comparison for altitude prediction results based on  3 different flights data;  (b): Altitude constraint violation comparison between constrained (454 violations) and unconstrained (675 violations)  model. }
\label{f:landing point constraint}
\end{figure}

From Table \ref{t:mse},
the overall accuracy of the prediction of both the 15-second sequence and the last landing point is higher after adding knowledge constraints. Baseline BNN model obtains the lower error in latitude and course predictions but those results are pretty close. Based on the knowledge of these two variables, they are both related to the direction and one flight is almost approaching along a straight line to the runway. Thus, the latitude and course data near the landing points is noisy and no models 
can gain significant advantage over the others. 

\begin{figure}[!t]
\centering
\includegraphics[width=.7\textwidth]{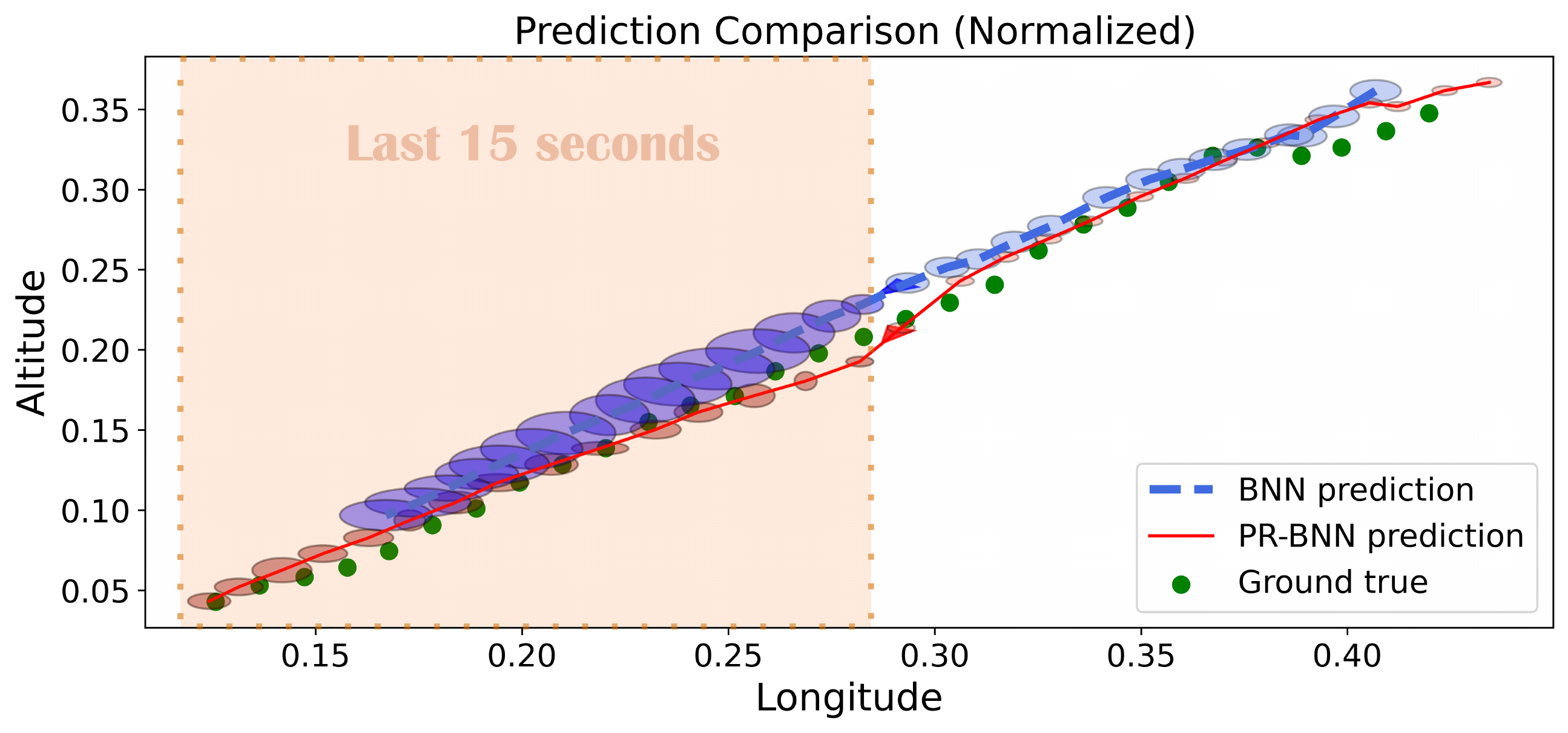}
\caption{Sample prediction results comparison with uncertainty: The shaded area shows the last 15-second predictions for two model and the ground true power output values.  The height and width of the markers are  proportional to the uncertainty of altitude and longitude predictions.}
\label{f:2d prediction results}
\end{figure}

From Table \ref{t:std}, an obvious reduction in the uncertainty of predictions is exhibited when we use the knowledge-constrained model. Table \ref{t:crps} demonstrate the probabilistic prediction performance improvement of our constrained framework via CRPS scores. One of our constraints is boundary constraint on the altitude of the last landing point, which is marked as the range in Figure \ref{f:landing point constraint} (a)  by two red lines of dashes. We can see the constraint violation of the green flight without knowledge constraints, while the prediction satisfies the constraint after applying the constraints. Figure \ref{f:landing point constraint} (b) shows the  reduction   of altitude constraint violation with knowledge. Since in last 15 second, most fluctuation of latitude is noise, we visualize the vertical 2D  coordinate system including altitude and longitude. Figure \ref{f:2d prediction results} shows the performance difference of BNN and PR-BNN models and the corresponding ground true values for this flight as approaching. 

\subsection{Photovoltaic Solar Energy Modelling \label{subsec: solar}}

In a solar power plant, accurate prediction of the solar \revision{Photovoltaic} (PV) plant power is important for power modeling, forecasting, and anomaly detection. 
In this case study, we use some ambient variables and  operation rules provided by professionals to explore the AC power generated in the plant. Generally, a  plant consists of multiple solar power inverters and each inverter has several modules. In this dataset, we explore one inverter in a solar plant using the proposed PR-BNN framework.

\begin{figure}[!t]
\centering
\includegraphics[width=.9\textwidth]{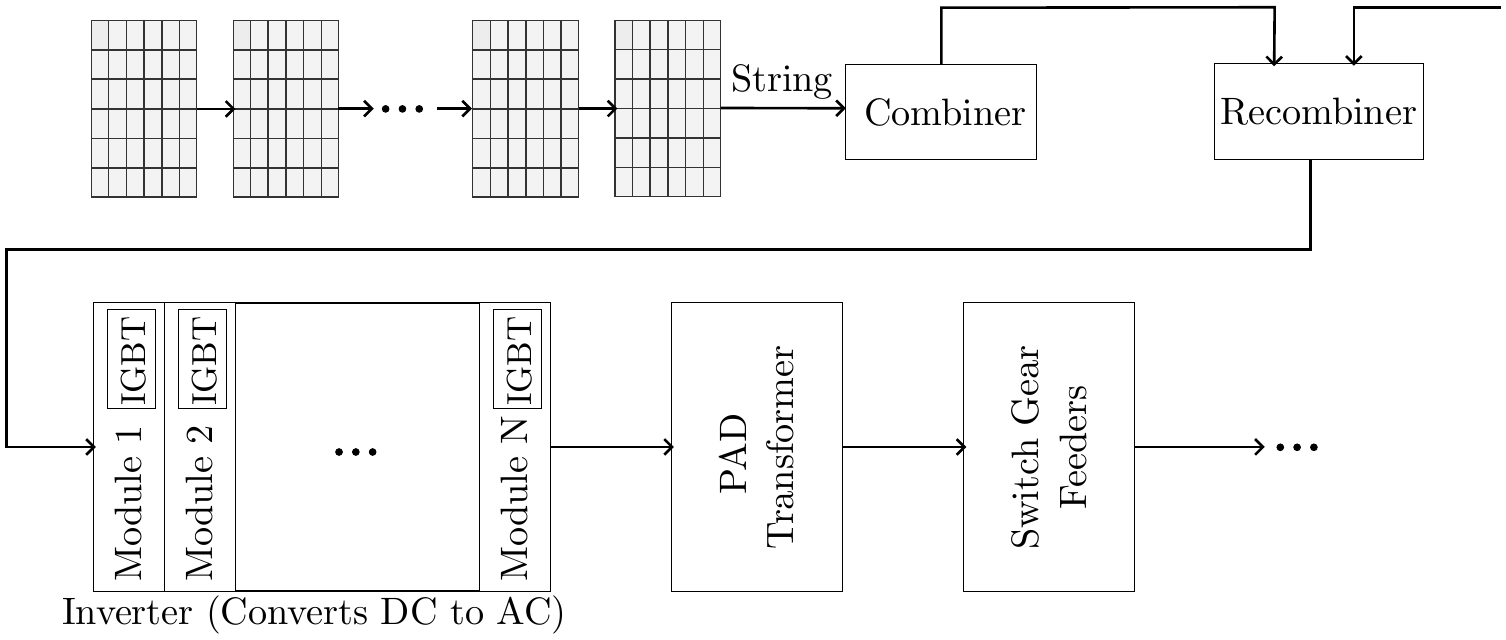}
\caption{Solar Energy System}
\label{f:plant}
\end{figure}

\revision{ Our data is provided by Arizona Public Service in the red rock PV plant. The core components in the solar energy system can be illustrated in Figure \ref{f:plant}. Various sensor data are available to measure the conditions at the system level and inverter level, and the weather station information is also provided on the site. }

In this dataset with the 15-minute resolution, our target variable is the AC power (W) of inverter A. \revision{The summer season is pretty long at the specific plant location in Arizona, and the extreme heat in summer often causes issues in this site, as experts described. Thus, we select only summer season data from May to October in 2019 and 2020. We have 20698 samples used in training and 6505 in testing.}
The independent variables are organized into two levels.  First, system-level variables contain  some control-related variables such as  curtailment, combiner switch, feeder switch, and some weather-related factors including wind speed, and irradiance. \revision{Second, the inverter-level data includes two binary variables related to IGBT (Insulated-Gate Bipolar Transistor) temperatures as shown in the second and third variables of Table \ref{t:solar variables}. IGBT is a core component of each module, and indicates the sometimes indicating the working condition of the corresponding module. We have some knowledge about the IGBT temperature. For example, if IGBT\_derate value is 1, it means that at least one module in inverter A has the derate condition due to the high IGBT temperature.} 
Daytime  and date variables are included to represent the temporal relationship in the simple BNN model.      \revision{All variables used in this case study are listed in Table \ref{t:solar variables}.}
\begin{table}[h]
\centering
\caption{Variables and description}
\revision{\begin{tabular}{|c|c|c|c|}
\hline
\textbf{Level}            & \textbf{Viarable} & \textbf{Description}                     & \textbf{Unit}          \\ \hline
\multirow{3}{*}{Inverter} & Pac                    & AC Power                                 & w                      \\ \cline{2-4} 
                          & IGBT\_derate           & $>$ IGBT temperature   derate threshold & -                      \\ \cline{2-4} 
                          & IGBT\_off              & $>$ IGBT temperature   off threshold    & -                      \\ \hline
\multirow{8}{*}{System}   & curtail\_utility       & Utility curtailment   power set point    & mw                     \\ \cline{2-4} 
                          & curtail\_on            & Curtailment OFF/ON                       & -                      \\ \cline{2-4} 
                          & P\_watts               & Main Xfmr3 Sec. power                    & mw                     \\ \cline{2-4} 
                          & Feeder\_B      & Feeder 2 Breaker                         & -                      \\ \cline{2-4} 
                          & WS                     & Wind Speed                               & m/s                    \\ \cline{2-4} 
                          & IPOA                   & Irradiance - plane of   array            & w/m\textasciicircum{}2 \\ \cline{2-4} 
                          & day\_label             & The day in a year                        & -                      \\ \cline{2-4} 
                          & time\_label            & The timestamp in a   day                 & -                      \\ \hline
\end{tabular}}
\label{t:solar variables}
\end{table}

In the data prepossessing, min-max scalar has been applied for both input and target variables for  normalization as shown in Equation \eqref{e:scaler}.

\begin{figure}[!t]
\centering
\includegraphics[width=0.85\columnwidth]{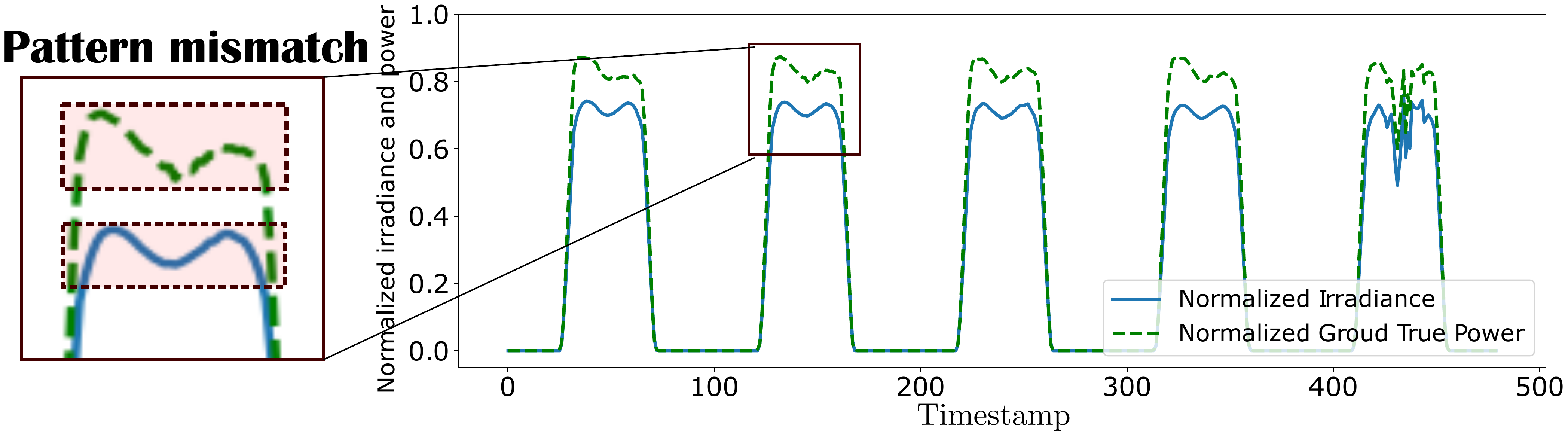}
\caption{Sample  variable visualization: \revision{x-axis is the time stamp; the y-axis represents the normalized irradiance value and the ground true AC power output values. The comparison of these two variables} shows pattern mismatch during the peak time or daytime on some days.}
\label{f:solar irradiance}
\end{figure}

\revision{In the solar energy plant, it is well known that solar irradiance has a direct impact on the power output, and it can explain most variability of the AC power output, as shown in Figure \ref{f:solar irradiance}. But for some variability, especially in the dashed rectangular areas, the solar irradiance alone cannot explain such variability. We would like to include other variables and even professional knowledge into the model to learn complex patterns.}
 
\begin{figure}
    \centering
    \includegraphics[width=0.8\columnwidth]{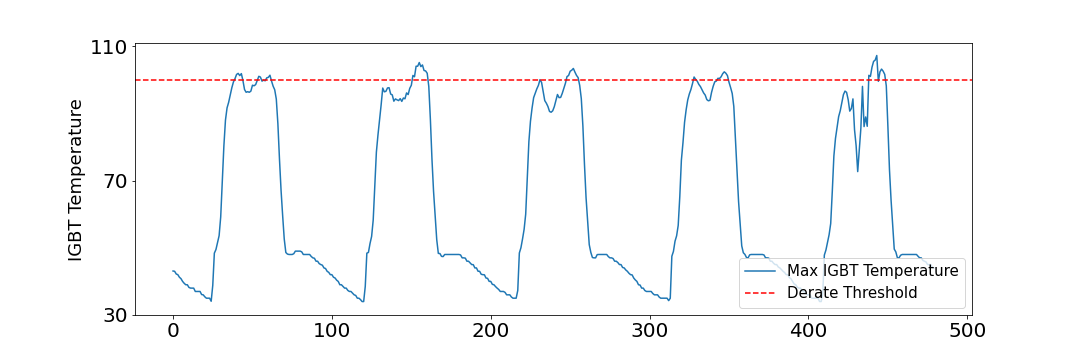}
    \caption{Derate condition based on \revision{Max} IGBT temperature}
    \label{fig:igbt}
\end{figure}

\revision{For the knowledge constraint, we follow the operational guide by APS that if the Max IGBT temperature reaches a threshold given by the expert, as shown in Figure \ref{fig:igbt}, the corresponding module will be derated to protect the machine so that IGBT temperature could be back to the safe value. As a result, the AC power will go down 
given that other conditions are unchanged, such as irradiance.} 

\begin{figure}
    \centering
    \includegraphics[width=0.6\columnwidth]{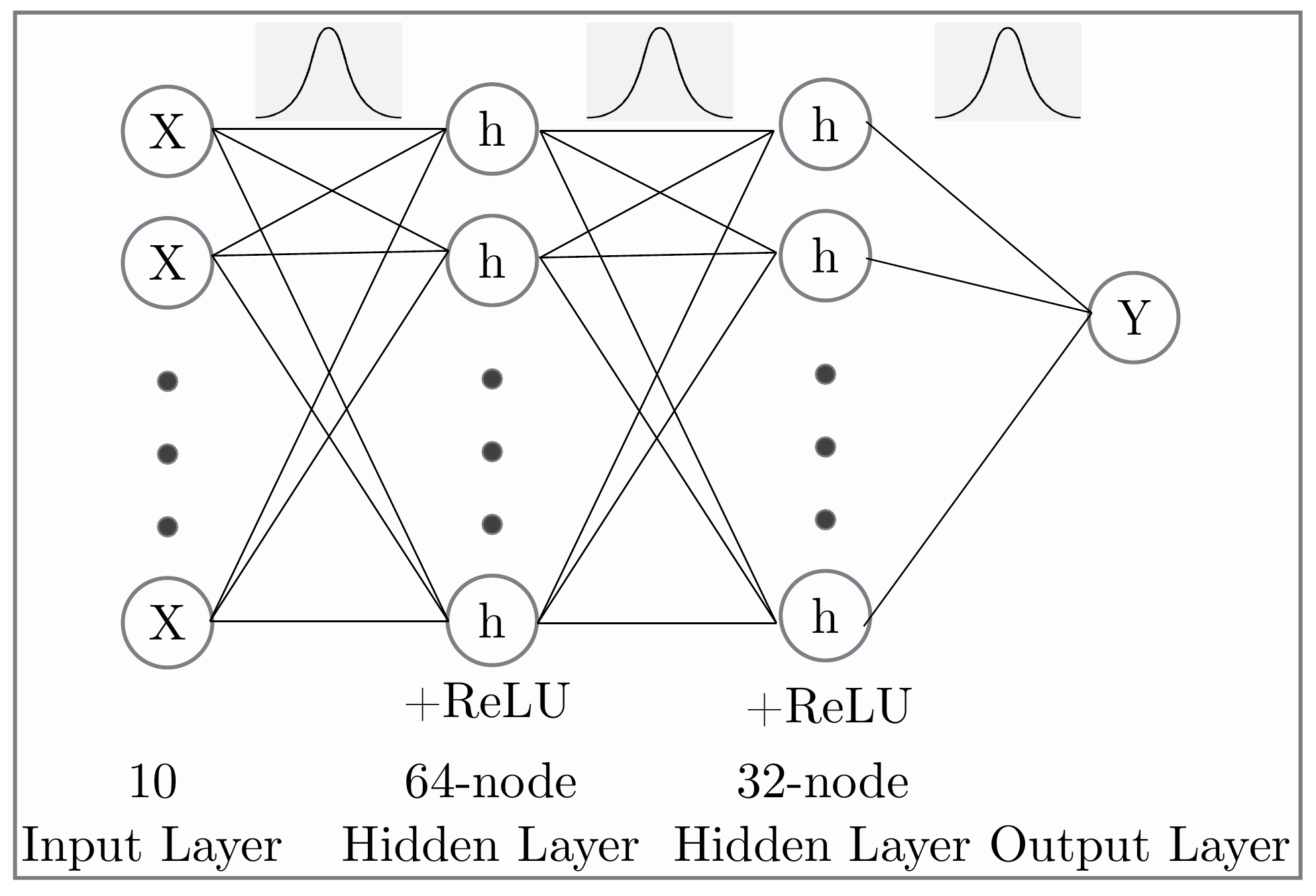}
    \caption{\revision{Baseline Bayesian Linear Neural Network Structure}}
    \label{f:solar_model}
\end{figure}

 \revision{A simple BNN with three Bayesian linear layers is applied as a benchmark \revision{BNN model as shown in Figure \ref{f:solar_model}}. The three layers have 64, 32, and 1 node, respectively, with Gaussian prior for Neural Network parameters, and the first two layers are followed by the ReLU activation function. }Based on the practical goals, the daytime (9:30 am - 5:00 pm) and peak time (10:00 am - 2:00 pm) performances are of most interest. Thus, we also use Mean Square Error (MSE), Root Mean Squared Error (RMSE) based on rescaled data to measure the mean prediction accuracy, and Continuous Ranked Probability Score (CRPS) based on rescaled data to evaluate the probabilistic forecasting quality. Evaluation comparisons between the Benchmark BNN model and the proposed PR-BNN model are shown in Table \ref{t:solar_evalution}. After adding the derate constraint, the MSE is reduced from 0.0018 to 0.0011. Performance improvements can be also observed by the RMSE and CRPS. 

\begin{table}[!htbp]
\centering
\caption{Performance comparison based on different time ranges}
\begin{tabular}{|c|ccc|ccc|}
\hline
Model & \multicolumn{3}{c|}{BNN}                                              & \multicolumn{3}{c|}{PR-BNN}                                           \\ \hline
Time  & \multicolumn{1}{c|}{All}    & \multicolumn{1}{c|}{Daytime} & Peaktime & \multicolumn{1}{c|}{All}    & \multicolumn{1}{c|}{Daytime} & Peaktime \\ \hline
MSE   & \multicolumn{1}{c|}{0.0018} & \multicolumn{1}{c|}{0.0045}  & 0.0037   & \multicolumn{1}{c|}{0.0011} & \multicolumn{1}{c|}{0.0026}  & 0.0017   \\ \hline
RMSE  & \multicolumn{1}{c|}{0.0419} & \multicolumn{1}{c|}{0.0672}  & 0.0611   & \multicolumn{1}{c|}{0.0327} & \multicolumn{1}{c|}{0.0505}  & 0.0415   \\ \hline
CRPS  & \multicolumn{1}{c|}{0.0186} & \multicolumn{1}{c|}{0.0448}  & 0.0450   & \multicolumn{1}{c|}{0.0114} & \multicolumn{1}{c|}{0.0233}  & 0.0212   \\ \hline
\end{tabular}
\label{t:solar_evalution}
\end{table} 

\begin{figure}[!t]
\centering
\includegraphics[width=1.05\columnwidth]{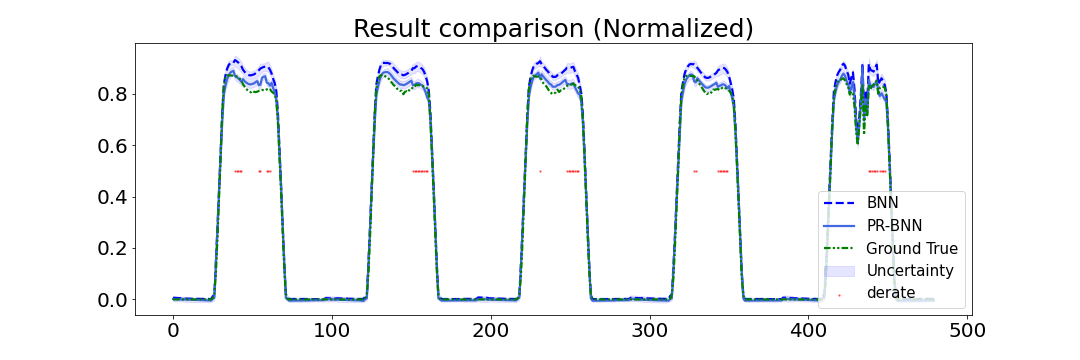}
\caption{Sample  variable visualization:
\revision{The x-axis is timestamp; The y-axis represents the normalized ground true and predictions of two models. The y-axis value of the red scatters no real meaning, but it indicates that there should be a manual derate at a certain timestamp with a red dot.}}
\label{f:solar result}
\end{figure}

Sample prediction results from both with and without constraint models for six days are visualized in Figure \ref{f:solar result}.  From the comparison, the PR-BNN model with derate-constraint learns the derate effect in the solar power system more accurately, which leads to improvement in overall accuracy in prediction.

 The review of the corresponding solar irradiance as shown in Figure \ref{f:solar irradiance} can make clear that the shape property of BNN baseline prediction is more similar to the irradiance curve shape but cannot learn the derate pattern from the data itself. Thus adding derate knowledge can help improve the accuracy of the regression and further may help detect the anomaly in the energy plant.

\section{Conclusion}
The performances of deep learning models can be benefited from rich domain knowledge or side information via BNNs, even for those cases with poor or limited datasets in HD problems. The PR-BNN with a general form of constraint functions based on posterior regularization is proposed. Due to the various properties of  domain knowledge, we include both types of soft and hard constraints in this work and develop efficient optimization algorithms for both cases. 


After implementing both soft and hard knowledge constraints with proposed algorithms, favorable results are demonstrated by our simulation results, showing promising potentials for performance and accuracy improvement. Besides simulation results, the proposed method is evaluated through two real-world case studies from different fields, namely trajectory prediction and solar energy modeling. Along with the accuracy metric, CRPS is adopted to evaluate and compare the probabilistic forecast performance of BNNs and PR-BNNs. The two case studies illustrate how to apply various professional or domain knowledge to real-world applications and show desirable performance on both deterministic and probabilistic prediction accuracy.

\section*{Acknowledgment}
The research reported in this paper was supported by funds from NASA University Leadership Initiative program (Contract No. NNX17AJ86A, PI: Yongming Liu, Technical Officer: Anupa Bajwa) and funds from Department of Energy (Contract No. DE-EE0009354, PI: Hao Yan). The support is gratefully acknowledged.

\bibliography{literature.bib}

\end{document}